\documentclass[a4paper,conference]{IEEEtran}

\usepackage{graphicx}
\usepackage{xcolor}[table]
\usepackage{amsmath}
\usepackage{amssymb}
\usepackage{multirow}
\usepackage{algorithm}
\usepackage{algcompatible}
\usepackage{dsfont}

\usepackage{caption}
\begin{document}
\title{Efficient Gaussian Process Model on Class-Imbalanced Datasets \\ for Generalized Zero-Shot Learning}
%\title{Inexpensive Gaussian Processes for Class-Imbalanced \\ Generalized Zero-Shot Learning}

\author{\IEEEauthorblockN{Changkun Ye}
\IEEEauthorblockA{
Australian National University \& Data61 CSIRO \\
Canberra, ACT, Australia\\
Email: changkun.ye@anu.edu.au}
\and
\IEEEauthorblockN{Nick Barnes}
\IEEEauthorblockA{Australian National University\\
Canberra, ACT, Australia\\
Email: nick.barnes@anu.edu.au}
\and
\IEEEauthorblockN{Lars Petersson and Russell Tsuchida}
\IEEEauthorblockA{Data61 CSIRO\\
Canberra, ACT, Australia\\
Email: lars.petersson@data61.csiro.au\\
russell.tsuchida@data61.csiro.au}
}

\maketitle

\begin{abstract}
Zero-Shot Learning (ZSL) models aim to classify object classes that are not seen during the training process. However, the problem of class imbalance is rarely discussed, despite its presence in several ZSL datasets. In this paper, we propose a Neural Network model that learns a latent feature embedding and a Gaussian Process (GP) regression model that predicts latent feature prototypes of unseen classes. A calibrated classifier is then constructed for ZSL and Generalized ZSL tasks. Our Neural Network model is trained efficiently with a simple training strategy that mitigates the impact of class-imbalanced training data. The model has an average training time of 5 minutes and can achieve state-of-the-art (SOTA) performance on imbalanced ZSL benchmark datasets like AWA2, AWA1 and APY, while having relatively good performance on the SUN and CUB datasets.
%\NB{Entire? Aren't their features from a base classifer there? - perhaps Using features pre-trained for image classification}
\end{abstract}
% Old version: Zero-Shot Learning (ZSL) aims to classify object classes that are not seen during the training process. In this paper, we proposed a novel Episode-based triplet loss training model to efficiently obtain separable image features for ZSL datasets and a Gaussian Process Regression model to predict prototypes of unseen classes. Our model have an average training time of 5 minutes while achieves state-of-the-art or comparable performance on several benchmard ZSL datasets.

%-------------------------------------------------------------------------
\section{Introduction}
\label{sec:intro}
{\it Zero-Shot Learning} (ZSL) requires a model to be trained on images that show examples from one set of classes, referred to as {\it seen classes}, while being tested on images that show examples from another set of classes, referred to as {\it unseen classes}. During training, semantic information for both seen classes and unseen classes is provided to help infer the appearance of unseen classes.

Many previous works, such as \cite{conse,SYNC,GFZSL,GDAN,f-CLSWGAN}, focus on learning a mapping between image features depicting certain classes and their corresponding semantic vectors. GFZSL \cite{GFZSL} proposed a model similar to Kernel Ridge Regression to predict image features of unseen classes. GDAN \cite{GDAN} and f-CLSWGAN \cite{f-CLSWGAN} utilize generative models like GAN \cite{GAN} and VAE \cite{VAE} to achieve the same objective.

On the basis of these approaches, recent papers further learn a Neural Network (NN) projection from image feature space to a latent embedding space, where inter-class features can be better separated within each ZSL dataset \cite{tripletloss,RFF,FREE,DVBE,CE-GZSL}. For example, in \cite{RFF}, image features are projected to a latent space in order to ``remove redundant information". FREE \cite{FREE} adopts the same structure for ``feature refinement" purposes. CE-GZSL \cite{CE-GZSL} also proposes a similar approach to generate a ``contrastive embedding" of image features.

\begin{figure}
	\centering
	\includegraphics[width=3.3in]{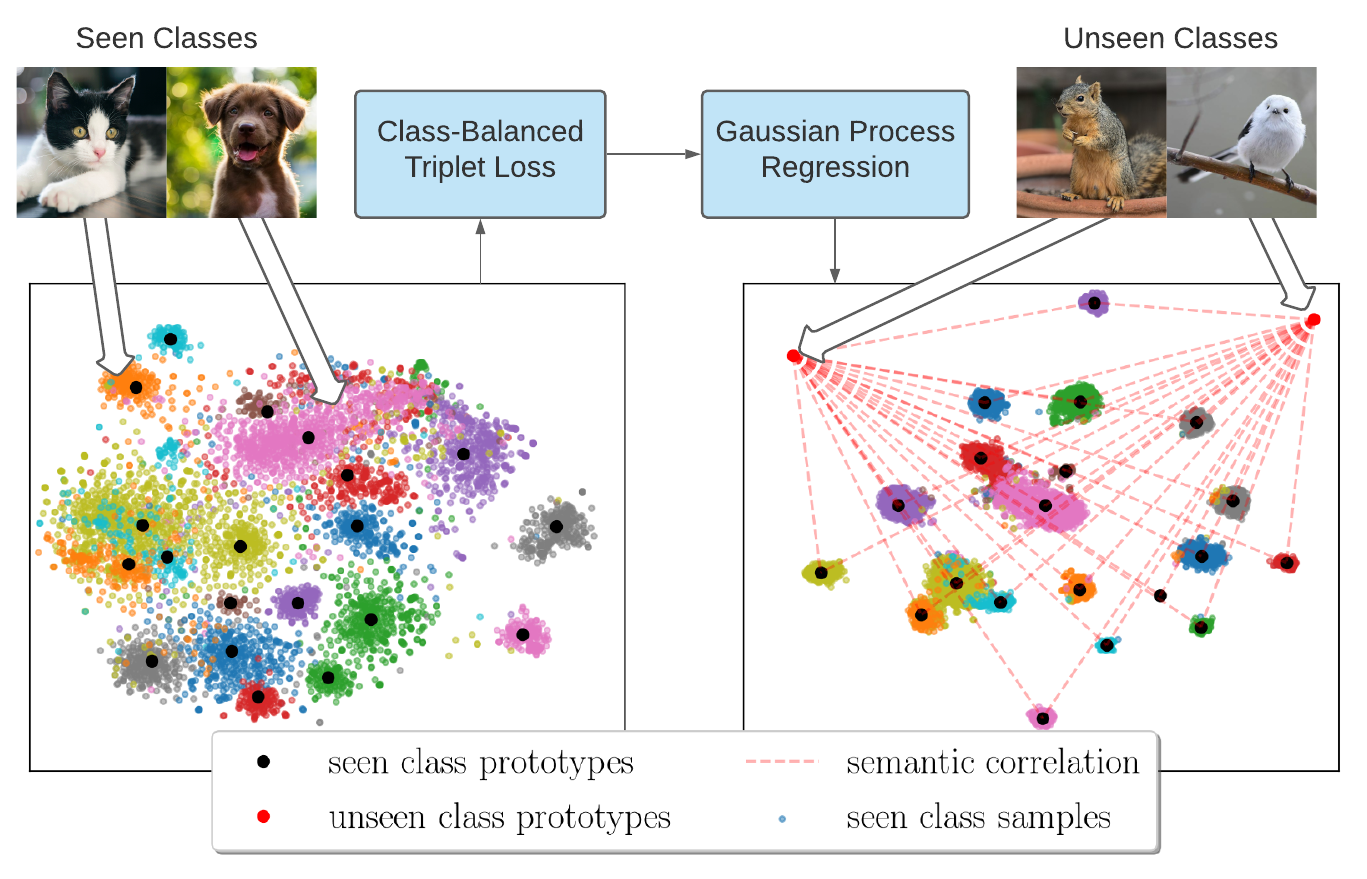}
	\label{Fig:11}
	\caption{We first train a latent embedding model for image features. The model is trained with Class-Balanced Triplet loss in order to separate inter-class features, which is robust to class-imbalanced datasets. Then a Gaussian Process Regression model is proposed to predict unseen class prototypes based on seen class prototypes and semantic correlations between classes. Finally, our ZSL classifier is constructed based on the prototypes.}
\end{figure}

Previous models, however, do not typically concern themselves with the class-imbalanced data distributions of ZSL datasets. In the real visual world, visual datasets usually exhibit an imbalanced data distribution among categories \cite{openlongtailrecognition}. In supervised learning, the class imbalance problem can have significant impact on the performance of classification models \cite{buda2018systematic,cui2019classbalancedloss}. For the ZSL problem, the APY \cite{APY} dataset has nearly $1/3$ of samples belonging to the same class. AWA2 \cite{res101} has 1645 samples in one class and only 100 samples in another. Clearly, the class imbalance problem is not negligible when training a classification model on these datasets.

On the other hand, recent models usually have complicated structures that require strong regularizers in order to prevent overfitting on seen class samples. As a consequence, these models usually have long training times and heavy GPU memory usage. The average training time for DVBE \cite{DVBE} is over 2 hours on each ZSL dataset. This fact motivates us to search for alternative models that are simpler and less prone to overfitting.

In this paper, we adopt the idea of projecting image features in a latent embedding space via a Neural Network (NN) model. We propose a class-balanced triplet loss that separate image features in a latent embedding space for class-imbalanced datasets. We also propose a Gaussian Process (GP) model to learn a mapping between features and a semantic space. The classical Gaussian Process (GP), when used in the setting of regression, is robust to overfitting \cite{gpbook}. If training and testing data come from the same distribution, a PAC-Bayesian Bound \cite{PAC2012} guarantees that the training error will be close to the testing error. %Several previous papers also proposed to use GP in the ZSL regime \cite{GPRZSL,GPRZSL2,GPRZSL3}, but none of them have a similar architecture to ours, nor do they have comparable performance.

Our experiments demonstrate that our model, though employing a simple design, can reach SOTA performance on the class-imbalanced ZSL datasets AWA1, AWA2 and APY in the Generalized ZSL setting.

The main contributions of our work are:
\begin{enumerate}
  \item We propose a novel simple framework for ZSL, where image features from a deep Neural Network are mapped into a latent embedding space to generate latent prototypes for each seen class by a novel triplet training model. Then a Gaussian Process (GP) regression model is trained via maximizing the marginal likelihood to predict latent prototypes of unseen classes. 
  \item The mapping from image features to a latent space is performed by our proposed triplet training model for ZSL learning, using a novel triplet loss that is robust on class-imbalanced ZSL datasets. Our experiments show improved performance over the traditional triplet loss on all ZSL datasets, including SOTA performance on class-imbalanced datasets, specifically, AWA1, AWA2 and APY. 
  %\NB{You can push model if you like, but is the loss itself novel? Is that the major difference?}
  \item Given feature vectors extracted by a pre-trained ResNet, our model has an average training time of 5 minutes on all ZSL datasets, faster than several SOTA models that have high accuracy. 
\end{enumerate}
%\NB{Urgently, you need to sort out your three contributions here. We need to get these carefully worded as this claim is essential to the paper}

\section{Related Works}
\textbf{Traditional and Generalized ZSL: }Early ZSL research adopts a so-called Traditional ZSL setting \cite{conse,ALE}. The Traditional ZSL requires the model to train on images of seen classes and semantic vectors of seen and unseen classes. Test images are restricted to the unseen classes. However, in practice, test images may also come from the seen classes \cite{res101}. The Generalized ZSL setting was proposed to address the problem of including both seen and unseen images in the test set. According to Xian {\it et al.}~ \cite{res101}, models that have good performance in the Traditional ZSL setting may not work well in the Generalized ZSL setting.

\textbf{Prototypical Methods.} Our classification model is related to prototypical methods proposed in Zero-Shot and Few-Shot learning \cite{Snell2017PrototypicalNF,prototype1,prototype2}. In the prototypical methods, a prototype is learned for each class to help classification. For example, Snell {\it et al. }  \cite{Snell2017PrototypicalNF} propose a neural network to learn a projection from semantic vectors to feature prototypes of each class. Test samples are classified via Nearest Neighbor among prototypes. While the classification process of our model is similar to prototypical methods, our model uses a Gaussian Process Regression instead of Neural Networks to predict prototypes of unseen classes. 

\textbf{Inductive and Transductive ZSL: } Similar to most ZSL models, the model we propose is an inductive ZSL model. Inductive ZSL requires that no feature information of unseen classes is present during the training phase \cite{res101}. Models that introduce unlabeled unseen images during the training phase are called transductive ZSL models \cite{transductive0}. Ensuring a fair comparison, results from such models are usually compared separately to inductive models since additional information is introduced \cite{GFZSL,rethinking,transductive1}. 

%These papers usually us GP to directly predict  \cite{DKL,SV-DKL,DNGP,GP2,GP3}\russ{Can you disambiguate this a bit more? I had a look at the first paper you cite, and could immediately determine what you are trying to say. Is the GP prior placed over the image, or a regression target over the space of images, or a classification target over the space of images?}.

\textbf{Triplet Loss.} %Triplet loss are usually proposed in Neural Network models\russ{word choice here is a little bit confusing. ``objective'' and ``loss'' are sometimes referring to the same thing. Can you replace the word objective with something else?} to help increase Euclidean distance between samples from different classes and reduce Euclidean distance between samples from the different class. 
Many ZSL models have proposed a triplet loss in their framework to help separate samples from different classes. Chacheux {\it et al.} \cite{tripletloss} proposed a variant of a triplet loss in their model to learn feature prototypes for different classes. Han {\it et al.} \cite{RFF} adopt an improved version of the triplet loss called ``center loss" proposed in \cite{centerloss} that separates samples in a latent space. Unlike their models, we notice that current triplet losses proposed for the ZSL problem may not perform well on class-imbalanced datasets like AWA2, AWA1 and APY. An improved version of the triplet loss training model is proposed to mitigate this problem.

\textbf{Gaussian Process Regression.} %Many research papers have developed Gaussian Process (GP) models for classification tasks. 
For the ZSL problem, Dolma {\it et al.} \cite{GPRZSL2} proposed a model that performs k-nearest neighbor search for test samples over training samples and performs a GP regression based on the search result. Mukherjee {\it et al.} \cite{GPRZSL3} model image features and semantic vectors for each class with Gaussians, and learns a linear projection between the two distributions. Our model is closest to Elhoseiny {\it et al.} \cite{GPRZSL}, where Gaussian Process Regression is used to predict unseen class prototypes based on seen class prototypes. However, they used a Gaussian Process directly without the benefit of a learned network model for feature embedding, and showed relatively poor results. Verma and Rai \cite{GFZSL} proposed a Kernel Ridge Regression (KRR) approach called GFZSL for the traditional ZSL problem. Our experiment demonstrates that our model outperforms GFZSL by a large margin. %Given that the hyperparameters of both models are suitably aligned, the posterior predictive mean of a GPR model is equal to the prediction made by KRR. 

%The Kernel Ridge Regression approach has similar form compare with Gaussian Process Regression. The practical difference is, hyperparameters in the KRR method is
%determined via grid search, whereas in GPR these hyperparameters can be trained by maximizing marginal likelihood. Our experiments demonstrate that the latter approach can give better performance. %\russ{For someone unfamiliar with GPR and KRR, it is not immediately clear what this sentence is saying. Can you help out a bit more by first explaining the difference and similarity between GPR and KRR? How do we choose hyperparameters in GPR, and how do we choose hyperparameters in KRR?}
%\NB{So there are three other GPR zero shot methods here. But you don't compare to any of them? Alsothe Elhoseiny that you say is closest you don't really offer any difference. What is the difference - just performance for us based on better innovations? Need more on this. Could specify our models.}
\begin{figure*}
	\centering
	\includegraphics[width=4.2in]{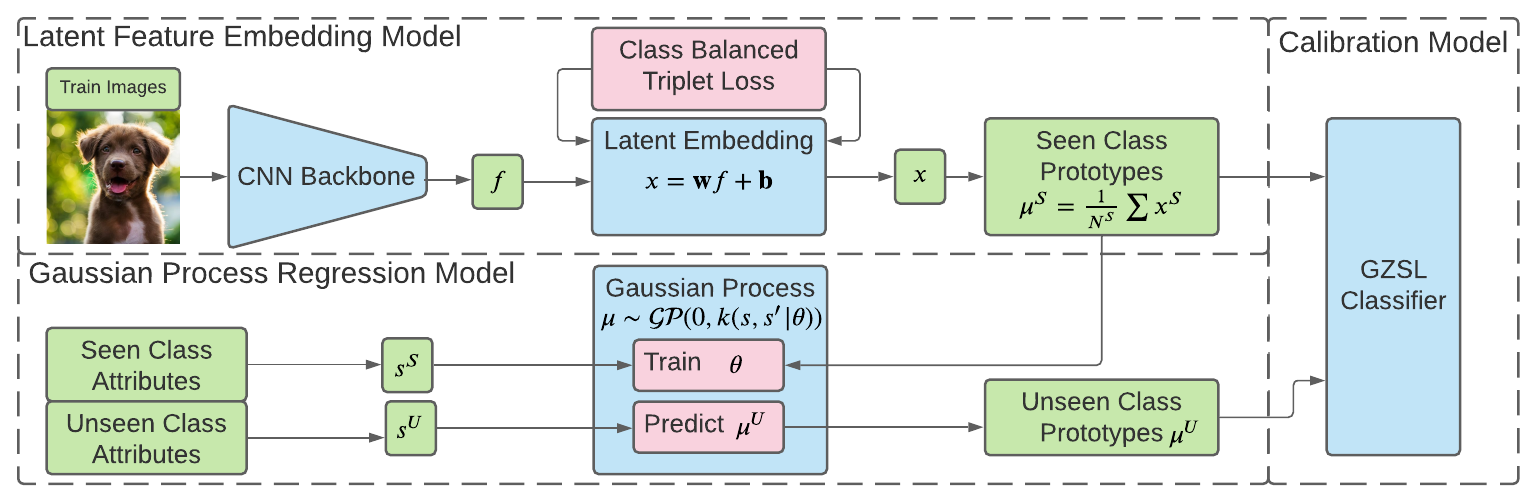}
	\caption{Structure of our proposed model. Feature vectors $\mathbf{f}$ are projected to a latent embedding space $\mathbf{x}$ which is trained using proposed Class-Balanced Triplet Loss. A GP model is proposed to predict latent prototypes of unseen classes $\mathbf{\mu}^U$, based on latent prototypes of seen class $\mathbf{\mu}^S$ and semantic vectors from seen and unseen class $\mathbf{s}^S,\mathbf{s}^U$.}
	\label{Fig:AUSUC}
\end{figure*}
\section{Proposed Approach}
%\subsection{Motivation} %In the ZSL setting, the feature vector $\mathbf{f}$ for each image- is extracted by a backbone network like ResNet \cite{RESNET} or VGG \cite{VGG}, pretrained on ImageNet \cite{ImageNet}.  If the features are not further adjusted with respect to each ZSL dataset, they may have cross-dataset bias \cite{CE-GZSL}, which can limit the performance for ZSL classification models.
%Real world visual datasets usually exhibit imbalanced data distribution among categories \cite{openlongtailrecognition}. This may have a significant impact on the performance of traditional classification model \cite{buda2018systematic}. In ZSL, however, class imbalance problem is rarely discussed, despite the fact that several ZSL datasets have class-imbalanced data distribution. This motivate us to tackle class imbalance problem in the ZSL setting.

We propose a hybrid model for the ZSL problem: a Latent Feature Embedding model to separate inter-class features that is robust to class-imbalanced datasets, a GP Regression model to predict prototypes of unseen classes based on seen classes and semantic information and a calibrated classifier to balance the trade-off between seen and unseen class accuracy.
%----------------------------Latent Space Generation------------------------------
\subsection{Latent Feature Embedding Model}
\textbf{Model Structure} We propose to learn a linear NN mapping from image features to latent embeddings. We argue that for the ZSL task, a linear projection with limited flexibility can help prevent the model from overfitting on seen class training samples. Following others \cite{GFZSL,GPRZSL2}, we model feature vectors from each class using the multivariate Gaussian distribution. We exploit the fact that Gaussian random vectors are closed under linear transformations.

For each feature vector $\mathbf{f}\in \mathbb{R}^{N_{feature}}$, the latent embedding $\mathbf{x}\in \mathbb{R}^{N_{latent}}$ can be written as:
\begin{equation}
    \mathbf{x} = \mathbf{wf} + \mathbf{b}.
\end{equation}
%\textcolor{red}{What is a feature vector? What is \mathbf{w}? what is $\mathbf{b}$?}
Here $\mathbf{w}\in \mathbb{R}^{N_{latent}\times N_{feature}}$ is a weight parameter matrix and $\mathbf{b}\in\mathbb{R}^{N_{latent}}$ is a bias parameter vector.

\textbf{Triplet Loss Revisited}
Triplet loss is often used to separate samples from different training classes in the dataset \cite{inbalancetriplet}. The standard triplet loss aims to decrease distances between intra-class samples and increase distances between inter-class samples. 

In each iteration, a mini-batch is sampled uniformly from training data as: $\{\mathbf{x}^{c_1}_{1},\mathbf{x}^{c_1}_{2},...,\mathbf{x}^{c_1}_{n^{c_1}},\mathbf{x}^{c_2}_{1},\mathbf{x}^{c_2}_{2},...\mathbf{x}^{c_2}_{n^{c_2}},...\mathbf{x}^{c_L}_{n^{c_L}}\}$. Here $\mathbf{x}^{c_i}_{j}$ denotes the $j^{th}$ our of total $n^{c_i}\in\mathbb{Z}^{+}$ samples that belongs to training class $c_i\in C, i= 1,2,...,L$ in mini-batch. The batch size is $N = \sum n^{c_j}$. Then all possible triplet pairs $\{\mathbf{x}^{c_i}_l,\mathbf{x}^{c_i}_m,\mathbf{x}^{c_j}_n\}$ are constructed within the given mini-batch. In each triplet, $\mathbf{x}^{c_i}_l,\mathbf{x}^{c_i}_m$ are different samples from the same class $c_i$, $l,m\in [1,2,...,n^{c_i}]$, $\mathbf{x}^{c_j}_n$ come from a different class $c_i\neq c_j$,  $n\in [1,2,...,n^{c_j}]$. The triplet loss is written as:
\begin{equation}
  \mathcal{L}_{T} = \sum_{c_i, c_j}\sum^{n^{c_i}}_{l=1}\sum^{n^{c_i}-1}_{m=1}\sum_{n=1}^{n^{c_j}}\max(0,\Delta + (\mathbf{x}_l^{c_i} - \mathbf{x}_m^{c_i})^2 - (\mathbf{x}_l^{c_i} - \mathbf{x}_n^{c_j})^2).
\end{equation}

%The $\max()$ operator helps filter out trivial triplets, where inter-class sample distance is already larger than the intra-class sample distance. 
The $\sum_{c_i,c_j}$ denotes summation over all training class pairs $c_i,c_j\in C$  that have $c_i\neq c_j$. Hyperparameter $\Delta\in\mathbb{R}^{+}$ is a positive threshold that balances the inter and intra class distances \cite{tripletloss,tripletbad1}. %\russ{This is a long sentence. Can you split it up into 2 or 3?} 
%\textcolor{red}{I am not convinced that $\Delta$ regularises the network. The first reference you cite makes the claim, but with no explanation or justification. Can you point me to where in the second reference it mentions regularisation via $\Delta$? Is it important to this paper that $\Delta$ performs a regularisation role?}

% \cite{tripletbad1} \russ{Can you clarify whether by \emph{convergence} you mean convergence of the sum, or convergence of an optimiser applied to the loss? \emph{What} is converging?}
The class imbalance problem is not considered in the original triplet loss. Moreover, models trained with a triplet loss usually require many iterations until convergence, expensive memory requirements and a high variance \cite{tripletbad1}. We thus propose a new Class-Balanced Triplet loss to mitigate these problems.

\textbf{Class-Balanced Triplet Loss}
When training a model with a triplet loss, a straight forward approach to tackle the class imbalance problem is to sample class-balanced mini-batch data. The model will not be affected by the class imbalance problem if it is trained using class-balanced data. 

In every iteration, unlike for the traditional triplet loss, we generate a class-balanced mini-batch by sampling $n^{CB}\in\mathbb{Z}^{+}$ data points from each one of $L$ training classes in the training set as  $\{\mathbf{x}^{c_1}_{1},\mathbf{x}^{c_1}_{2},...,\mathbf{x}^{c_1}_{n^{CB}},\mathbf{x}^{c_2}_{1},\mathbf{x}^{c_2}_{2},...\mathbf{x}^{c_2}_{n^{CB}},...\mathbf{x}^{c_L}_{n^{CB}}\}$. The batch size becomes $N=n^{CB}\times L$. In a supervised classification setting, similar approaches have shown to be effective \cite{buda2018systematic}.

We then propose a modified triplet loss $L_{BT}$ to train the model on the mini-batch. For every mini-batch, the loss has the form:
\begin{equation}
  \mathcal{L}_{BT} = \sum_{c_i,c_j}\sum_{l=1}^{n^{CB}}\max(0,\Delta + (\mathbf{x}_l^{c_i} - \overline{\mathbf{x}^{c_i}})^2 - \min_n(\mathbf{x}_n^{c_j} - \overline{\mathbf{x}^{c_i}})^2).
\end{equation}
The term $\overline{\mathbf{x}^{c_i}} = \frac{1}{n^c}\sum_n \mathbf{x}^{c_i}_n$ denotes the average of samples from class $c_i$ in the mini-batch. Replacing the term $\mathbf{x}_m^{c_i}$ with $\overline{\mathbf{x}^{c_i}}$ in the original triplet loss can help reduce the variance in the loss during training, which is similar to ``center loss" \cite{centerloss}. However, unlike their method, we are not adding extra trainable parameters into the model. The $\min()$ operation is performed over all samples $\mathbf{x}^{c_j}_m$ in class $c_j$ in the mini-batch, which can efficiently reduce computational costs. %This triplet loss model only needs to sum over the number of samples $n$ in class $c_i$ rather than over $l,m$ and $n$. 
%\NB{I'm not sure you've sufficiently explained this. With (3) What is N? you haven't defined it. Is N over classes? or over the class in the mini-batch. You say its average of class i in mini-batch, but N doesn't suggest that as a notation. I don't think you've defined the $x_l^{c_i}$ notation. I would assume l indexes an instance of class $c_i$ but I don't think this is clear.Balanced classes for mini-batch - do you choose just two classes at a time?}

%\russ{What is a separable feature?}
With the help of the proposed triplet loss $\mathcal{L}_{BT}$, our model can efficiently learn a latent embedding that separates samples from different classes and maintains a good performance on imbalanced datasets.

\subsection{Gaussian Process (GP) Regression Model}
%\russ{Is it possible to change this section as we discussed?}
%\russ{Love this update!}
We propose a GP Regression model to predict prototypes of unseen classes{, leveraging the generalization ability of GP models.} Like Mukherjee {\it et al.} \cite{GPRZSL3}, we obtain the average of all latent features in each class $\mathbf{\mu}^{c_i} = \frac{1}{N^{c_i}}\sum \mathbf{x}^{c_i}$ as a prototype for the corresponding class. 

%we model image features of each class with a multivariate Gaussian distribution. Since \NB{Gaussianity is} \sout{Gaussianness} preserve\NB{d} for linear operations, it is also plausible to model latent embedding of feature vector with a Gaussian: Parameter\NB{s} $\mathbf{\mu}^c$ are treated as prototypes for each class, which can be estimated using Maximum Likelihood Estimation: $\mathcal{N}(\mathbf{\mu}^{c},\Sigma^{c}), c\in C$.
%\begin{equation}
%  \mu^{c} \approx \overline{\mathbf{x^{c}}}
%\end{equation}
%\textcolor{red}{Perhaps I missed it, but I can't see where the semantic vectors' shapes are defined? Same for the various subscripts and superscripts of $\mu$.}
We also denote semantic vector of each class $s^{c_i}\in\mathbb{R}^{N_{semantic}}$. Given the semantic vectors $\mathbf{s}^S = [\mathbf{s}^{c_1},...\mathbf{s}^{c_L}]^T$ and feature prototypes $\mathbf{\mu}^{S} = [\mathbf{\mu}^{c_1},...\mathbf{\mu}^{c_L}]^T$ for seen class $c_1,c_2,...c_L \in C^S$, along with semantic vectors $\mathbf{s}^U=[\mathbf{s}^{c_{L+1}},...\mathbf{s}^{c_{L+K}}]^T$ for unseen classes $c_{L+1},c_{L+2},...c_{L+K} \in C^U$, we can use the GPR model to regress prototypes $\mu^{U} = [\mathbf{\mu}^{c_{L+1}},...\mathbf{\mu}^{c_{L+K}}]^T$ for unseen classes $c\in C^U$:
\begin{equation}
  \mathbf{\mu}^{U} = f_{GP}(\mathbf{s}^{U}|\theta) + \epsilon.
\end{equation}
%\NB{Mean is m, then you compute $\mu$? Your notation here is a bit confusing.}
Here $f_{GP}(\mathbf{s}|\theta)$ is the regression function, $\epsilon\sim\mathcal{N}(0,\sigma^2)$ denotes the Gaussian random noise and $\theta$ is the hyperparameter in the model. $\theta$ is trained given seen class semantic vectors $\mathbf{s}^S$ and corresponding prototypes $\mu^S$.

Directly training a GPR model that learns a projection from $\mathbf{s}$ to $\mathbf{\mu}$ requires accounting for every dimension in $\mathbf{\mu}\in\mathbb{R}^{N_{latent}}$, which is computationally expensive because the model needs to estimate correlations between different dimensions. We propose to avoid this issue by assuming dimensions in $\mathbf{\mu}^c$ are independent from each other so that the GPR model can be applied to $\mathbf{\mu}_i$:
\begin{equation}
  \mathbf{\mu}^U_i = f_{GP}(\mathbf{s}^U|\theta_i) + \epsilon_i.
\end{equation}
Then we have hyperparameter $\theta_i$ and noise $\epsilon_i$ for $i^{th}$ dimension in $\mu$.

A Gaussian Process is defined by a mean function $m(\mathbf{s})$ and a covariance function $k(\mathbf{s},\mathbf{s}'|\theta)$ that depends on hyperparameter $\theta$. For $f_{GP}(\mathbf{s}|\theta_i)$, a GP can be written as:
\begin{equation}
  f_{GP}(\mathbf{s}|\theta_i) \sim \mathcal{GP}(\mathbf{m}(\mathbf{s}),k(\mathbf{s},\mathbf{s}'|\theta_i)).
\end{equation}
Here we will take $\mathbf{m} \equiv 0$. The joint prior distribution of seen class prototypes $\mathbf{\mu}_i^S$ and regression function $f_{GP}(\mathbf{s}|\theta_i)$ can be written as:
\begin{equation}
  \begin{bmatrix}f_{GP}\\{\mathbf {\mu}}_i^{S}\end{bmatrix} \sim
	\mathcal{N}\left(\mathbf{0},
			\begin{bmatrix}
 			k(\mathbf{s},\mathbf{s}|\theta_i) & k(\mathbf{s},\mathbf{s}^S|\theta_i) \\
 			k(\mathbf{s}^S,\mathbf{s}|\theta_i) & k(\mathbf{s}^S,\mathbf{s}^S|\theta_i)+ \mathbf{I}\sigma_i^2 
			\end{bmatrix}\right),
\end{equation}
where $\mathbf{I}$ denotes the identity matrix. $f_{GP} = f_{GP}(\mathbf{s}|\theta_i)$ can be obtained via conditioning the joint prior distribution on $\mathbf{\mu}^S_i$ to obtain the posterior predictive distribution, which is also a Gaussian distribution $f_{GP}(\mathbf{s}|\theta_i)|\mathbf{\mu}_i^S,\mathbf{s}^S \sim \mathcal{N}(\mathbf{m}_i,\Sigma)$. We use the mean $\mathbf{m}_i$ of the predictive posterior distribution to form our prediction of $f_{GP}(\mathbf{s}|\theta_i)$ evaluated at the unseen classes $\mathbf{s}=\mathbf{s}^U$, which gives:
\begin{equation}
		f_{GP}(\mathbf{s}|\theta_i) = \mathbf{m}_i(\mathbf{s}) = k(\mathbf{s},\mathbf{s}^S|\theta_i) \left[ k(\mathbf{s}^S,\mathbf{s}^S|\theta_i) + \mathbf{I}\sigma_i^2\right]^{-1}\mu_i^S .
	\label{Eq:GP_covariance}
\end{equation}

Any positive semi-definite kernel function may be used as a covariance function $k(\mathbf{s},\mathbf{s}'|\theta_i)$, with $\theta_i$ as a hyperparameter in the kernel. We propose to search for optimal hyperparameters $\theta_i$ for each feature dimension $i$ by maximizing the log marginal likelihoods:

\begin{equation}
	\begin{aligned}
    \theta_i = \arg\max_{\theta_i} &\Bigg( -\frac{1}{2}(\mathbf{\mu}_i^S)^T\left[k(\mathbf{s}^S,\mathbf{s}^S|\theta_i)+ \mathbf{I}\sigma_i^2\right]^{-1}\mathbf{\mu}_i^S \Bigg.\\
    & \Bigg. -\frac{1}{2}\log k(\mathbf{s}^S,\mathbf{s}^S|\theta_i)\Bigg) .
  \end{aligned}	
\end{equation}
With $\theta_i$ given, unseen class prototypes can be evaluate by $\mu_i^U = f_{GP}(\mathbf{s}|\theta_i)$. Similar to other prototypical methods, the classifier can then be constructed using a nearest neighbor approach based on a distance metric. We use the Euclidean distance in our model:
\begin{equation}
  predict(\mathbf{x}) = \arg\min_{c\in C} ||\mathbf{x} - \mu^c||^2.
  \label{Eq:upredict}
\end{equation}
% RT
% equation (6) is by far more interesting than (2), (3), (4), (5). (2-5) are standard GPR stuff, but (6) is slightly different to the usual setting. I would recommend writing some more about (6) with the extra space you have from removing (2-5).
\subsection{Calibration}
It is well known that ZSL models trained on seen classes are inclined to be biased towards classifying unseen images into seen classes \cite{calibration,CNZSL}. Therefore, it is necessary to add a penalty term $\gamma\in\mathbb{R}^{+}$ when computing classification metrics over seen classes $c_i\in C^S$. We adopt the calibration approach proposed by Cacheux {\it et al.} \cite{calibration}. The calibrated nearest neighbor classifier is then written as:
\begin{equation}
  predict(\mathbf{x}) = \arg\max_{c\in C}[-||\mathbf{x}-\mathbf{\mu}^{c}||^2 - \gamma\mathds{1}_{c\in C^S}].
\end{equation}
where $\mathds{1}_{c_i\in C_S}$ is the indicator function, which equals to 1 when class $c$ is from a seen class and 0 otherwise. 

In our model, we first use the GPR model to predict the validation class based on the training class, then we train $\gamma$ as a calibration penalty to maximize the harmonic mean. After training, the test class is predicted and conditioned on training classes and validation classes together, and $\gamma$ is used for calibration evaluation of the performance on the test set. 

\section{Experiments}
\textbf{Datasets:} We test the performance of our model on five benchmark datasets, namely:
Animals with Attributes 1 (AWA1) \cite{res101}, Animals with Attributes 2 (AWA2) \cite{res101}, A Pascal and A Yahoo (APY) \cite{APY},
Caltech UCSD Birds 200-2011 (CUB) \cite{CUB} and SUN Attribute (SUN) \cite{SUN}. 
%\NB{Put your emphasis on the class imbalanced as that is your problem. Your claim is that it boosts performance on class-imbalanced, while still being OK on class balanced. List and push the class imbalanced first so the reader comes in with correct expectations} 
%CUB is a fine-grained dataset containing different bird species. SUN is also a fine-grained scene image dataset. AWA2 and AWA1 are coarse-grained datasets containing different animals. APY is also a coarse-grained dataset of common objects. Each of the datasets provides attribute vectors to represent semantic information. 
Detailed information of the datasets used is provided in Table~\ref{Table:Dataset} below.

We also provide the total, average, maximum and minimum sample number per-class in Table \ref{Table:Dataset}. We can see that CUB and SUN are relatively class-balanced datasets because of a low variance in the number of samples per-class. While AWA2, AWA1 and APY are class-imbalanced datasets. In particular, for APY, one single class contains $1/3$ of the total number of samples in the dataset.

\begin{table}[ht]
  \centering
  %\footnotesize
  \scriptsize
  \begin{tabular}{|c|c|c|c|c|c|c|}\hline
     \multicolumn{2}{|c|}{Dataset}& CUB & SUN & AWA2 & AWA1 & APY\\ \hline
     \multirow{2}{*}{Class} & seen & 150 & 645 & 40 & 40 & 20\\ 
     \cline{2-7} & unseen & 50 & 72 & 10 & 10 & 12\\ \hline
     \multicolumn{2}{|c|}{Feature Dim} & 2048 & 2048 & 2048 & 2048 & 2048\\ \hline
     \multicolumn{2}{|c|}{Attribute Dim} & 312 & 102 & 85 & 85 & 64\\ \hline \hline
     \multicolumn{2}{|c|}{Total sample No.} & 11788 & 14340 & 37322 & 30475 & 15339\\ \hline
     \multicolumn{2}{|c|}{Average Sample No. per-class} & 58 & 20 & 746 & 609 & 479 \\ \hline
     \multicolumn{2}{|c|}{Max Sample No. per-class} &60 & 20 & 1645 & 1168 & 5071 \\\hline
     \multicolumn{2}{|c|}{Min Sample No. per-class} & 41 & 20 & 100 & 92 & 51 \\\hline
  \end{tabular}
  \caption{Zero-Shot Learning Dataset Information. AWA1, AWA2 and APY are class-imbalanced datasets.}
  \label{Table:Dataset}
\end{table}
\textbf{``Proposed Split V2.0":} In the ZSL setting, datasets are usually divided into sets of seen classes and sets of unseen classes. Most recent models adopt ``Proposed Split" proposed by Xian {\it et al.} \cite{res101} to test the performance of their model.
%Because ZSL models usually use pre-trained CNNs like ResNet \cite{RESNET} to extract features, some classes in the ZSL dataset are present in the training process of those pre-trained CNNs. As Xian {\it et al.} \cite{res101} argues, such classes should not be put into the set of unseen classes when splitting the dataset for ZSL tasks. They further introduce a new split called ``Proposed Splits" to avoid this problem. The old split used by earlier ZSL models are then called ``Standard Splits".
On September 2020, Xian {\it et al.} \cite{res101} updated their paper with ``Proposed Split V2.0", in order to fix the problem that the old ``Proposed Split" has testing seen class samples included in the training samples. Such issues may have a big impact on current SOTA models' {performance}. In this work, we report performance of previous models reproduced on ``Proposed Split V2.0" by other papers as well as our own to ensure a fair comparison.

%\subsection{Traditional and Generalized ZSL setting}
%Earlier works on ZSL problems like \cite{DAPIAP,ALE} adopt the so-called Traditional ZSL setting. That is, when testing the performance, the model only classifies sample images from the set of unseen classes. In practice, however, the model is more likely to be required to classify images belonging to both the sets of seen and unseen classes, as argued by \cite{res101}. This is called the Generalized ZSL setting. Current papers such as \cite{GDAN,calibration,RFF,DAZLE,EPGN} focus more on the latter setting. Here, we test our model on both settings.
\begin{table*}[t]
\centering
\scriptsize
%\tiny
\setlength\tabcolsep{3pt}
\begin{tabular}{c | c||c|c c c||c|c c c||c|c c c||c|c c c ||c|c c c} 
 \multicolumn{21}{c}{ } \\ \hline
 \multirow{3}{*}{Methods} & \multirow{3}{*}{Provided by} & \multicolumn{4}{c||}{CUB} & \multicolumn{4}{c||}{SUN} & \multicolumn{4}{c||}{AWA2} & \multicolumn{4}{c||}{AWA1} & \multicolumn{4}{c}{APY} \\ \cline{3-22}
    & & ZSL & \multicolumn{3}{|c||}{GZSL} & ZSL & \multicolumn{3}{|c||}{GZSL} & ZSL & \multicolumn{3}{|c||}{GZSL} & ZSL & \multicolumn{3}{|c||}{GZSL} & ZSL & \multicolumn{3}{|c}{GZSL}\\ \cline{3-22}
    & & $A_T$ & $A_{U}$ & $A_{S} $ & $H$ & $A_T$ & $A_{U}$ & $A_{S}$ & $H$ & $A_T$ & $A_{U}$ & $A_{S}$ & $H$ & $A_T$ & $A_{U}$ & $A_{S}$ & $H$ & $A_T$ & $A_{U}$ & $A_{S}$ & $H$\\ \hline
\multicolumn{2}{c||}{ } & \multicolumn{8}{c||}{Class-Balanced Dataset} & \multicolumn{12}{c}{Class-Imbalanced Dataset} \\ \hline
SYNC \cite{SYNC} & \cite{res101} & 56.0 & 11.5 & 70.9 & 19.8 & 56.2& 7.9 & 43.3 & 13.4 & 49.3 & 9.7 & \textbf{89.7} & 17.5 & 51.8 & 9.0 & \textbf{88.9} & 16.3 &23.9 & 7.4 & 66.3 & 13.3 \\
ALE \cite{ALE} & \cite{res101} & 54.9 & 23.7 & 62.8 & 34.4 & 58.1 & 21.8 & 33.1 & 26.3 & 62.5 & 14.0 & 81.8 & 23.9 & 59.9 & 16.8 & 76.1 & 27.5 & 39.7 & 4.6 & 73.7 & 8.7 \\
DEVISE \cite{DEVISE}& \cite{res101} & 52.0 & 23.8 & 53.0 & 32.8 & 56.5 & 16.9 & 27.4 & 20.9 & 59.7 & 17.1 & 74.7 & 27.8 & 54.2 & 13.4 & 68.7 & 22.4 & 37.0 & 3.5 & 78.4 & 6.7 \\ 
GFZSL \cite{GFZSL}& \cite{res101} & 49.3 & 0.0 & 45.7 & 0.0 & 60.8 & 0.0 & 39.6 & 0.0 & 63.8 & 2.5 & 80.1 & 4.8 & 68.2 & 1.8 & 80.3 & 3.5 & 38.4 & 0.0 & \textbf{83.3} & 0.0 \\
GDAN \cite{GDAN} & \cite{CFZSL} & - & 35.0 & 28.7 & 31.6 & - & 38.2 & 19.8 & 26.1 & - & 26.0 & 78.5 & 39.1 & - & - & - & - & - & 29.0 & 63.7 & 39.9 \\
CADA-VAE \cite{CADA-VAE} & \cite{CFZSL}& - & 50.3 & 56.1 & 53.0 & - & 43.6 & 36.4 & 39.7 & - & 55.4 & 76.1 & 64.0 & - & - & - & - & - & 34.0 & 54.2 & 41.7 \\
TF-VAEGAN \cite{TF-VAEGAN} & \cite{CFZSL}& - & 50.7 & \textbf{62.5} & 56.0 & - & 41.0 & 39.1 & 41.0 & - & 52.5 & 82.4 & 64.1 & - & - & - & - & - & 31.7 & 61.5 & 41.8 \\ 
LisGAN \cite{LisGAN} & \cite{CFZSL}& - & 44.9 & 59.3 & 51.1 & - & 41.9 & 37.8 & 39.8 & - & 53.1 & 68.8 & 60.0 & - & - & - & - & - & 33.2 & 56.9 & 41.9 \\ 
%GDAN \cite{GDAN} & - & 14.4 & 22.4 & 17.5 & - & 5.3 & 23.2 & 8.6 & - & 19.2 & 58.9 & 28.9 & - & - & - & - & - & 17.4 & 57.3 & 26.7 \\
%CADA-VAE \cite{CADA-VAE} & author & - & \textbf{51.1} & 52.9 & 52.0 & - & 46.0 & 36.0 & 40.4 & - & 55.9 & 75.7 & 64.3 & - & 57.2 & 74.5 & 64.7 & - & - & - & - \\
Li {\it et al.} \cite{rethinking} & author & - & - & - & - & - & - & - & - & - & - & - & - & 69.4 & 59.2 & 78.4 & 67.5 & - & - & - & - \\
E-PGN \cite{EPGN}& author & \textbf{69.1} & 50.1 & 60.0 & 54.6 & - & - & - & - & 67.4 & 32.1 & 66.6 & 43.3 & \textbf{71.1} & 56.8 & 81.2 & 66.9 & - & - & - & -\\ 
DVBE \cite{DVBE} & author & - & 46.7 & 51.4 & 48.9 & - & 34.7 & 32.3 & 33.4 & - & 45.4 & 76.9 & 57.1 & - & - & - & - & - & 32.9 & 47.6 & 38.9 \\ 
GCM-CF \cite{CFZSL} & \cite{CFZSL}& - & \textbf{61.0} & 59.7 & \textbf{60.3} & - & 47.9 & 37.8 & 42.2 & - & 60.4 & 75.1 & 67.0 & - & - & - & - & - & 37.1 & 56.8 & 44.9 \\
CNZSL \cite{CNZSL} & \cite{CNZSL}& - & 49.9 & 50.7 & 50.3 & - & 44.7 & \textbf{41.6} & \textbf{43.1} & - & 60.2 & 77.1 & 67.6 & - & 63.1 & 73.4 & 67.8 & - & - & - & - \\ 
FREE \cite{FREE} & \cite{FREE} & - & 55.7 & 59.9 & 57.7 & - & 47.4 & 37.2 & 41.7 & - & 60.4 & 75.4 & 67.1 & - & 62.9 & 69.4 & 66.0 & - & - & - & - \\
%CNZSL \cite{CNZSL} & author & 54.5 & 49.9 & 50.7 & 50.3 & 60.8 & 44.7 & 41.6 & 43.1 & 68.1 & 60.2 & 77.1 & 67.6 & 69.7 & 63.1 & 73.4 & 67.8 & 37.3 & 29.4 & 57.6 & 38.9 \\\hline
%\multirow{2}{*}{RFF-GZSL \cite{RFF}5-nn }  & - & 48.9 & \textbf{78.8} & \textbf{59.6} & - & 49.5 & 42.4 & \textbf{45.7} & - & - & - & - & - & 54.2 & \textbf{94.3} & 68.4 & - & - & - & -\\ 
%& & $\pm$7.3 & $\pm$7.6 & $\pm$6.2 & & $\pm$4.5 & $\pm$1.4 & $\pm$1.8 & & & & & & $\pm$6.8 & $\pm$0.9 & $\pm$5.4 & & & & \\ \hline \hline 
\hline 
$\mathcal{L}_{BT}$ + GP(\textbf{ours}) & - & 59.9 & 50.1 & 56.3 & 53.1 & \textbf{63.2} & \textbf{50.4} & 34.8 & 41.2 & \textbf{68.6} & \textbf{62.2} & 76.7 & \textbf{68.7} & 70.1 & \textbf{64.5} & 73.3 & \textbf{68.6} & \textbf{47.1} & \textbf{42.8} & 64.3 & \textbf{51.4} \\ 
\hline
\end{tabular}
\caption{ZSL Top-1 per-class Accuracy on ``Proposed Split V2.0", Traditional ZSL as $A_T$, Generalized unseen, seen and harmonic mean as $A_U,A_S,H$ respectively. Our model clearly outperforms previous models on class-imbalanced datasets AWA2, AWA1 and APY.}
%\href{https://drive.google.com/file/d/1p9gtkuHCCCyjkyezSarCw-1siCSXUykH/view?usp=sharing}{Proposed Split V2.0}.
\setlength\tabcolsep{6pt}
\label{Table:performance}
\end{table*}

\textbf{Implementation Detail:} Our model is implemented using PyTorch and GPytorch \cite{gpytorch} and trained on an NVIDIA RTX 2080Ti GPU machine. We use feature vectors extracted by a pre-trained ResNet101 network, proposed by Xian {\it et al.} \cite{res101}. As argued by Chacheux {\it et al.} \cite{tripletloss}, the feature vector space is unbounded and a few feature vectors that have high values may hinder the network learning from the triplet loss. In this work we preprocess the feature vectors by clipping the features by $7$ and scaling to range $[0,1]$. The neural network model is trained with the Adam \cite{Adam} optimizer with learning rate $0.002$ and weight decay $0.1$ for 500 episodes on each ZSL dataset. We set the threshold $\Delta=4$ in our triplet loss. The GPR model is trained also with the Adam \cite{Adam} optimizer with learning rate $0.01$ for $1000$ epochs for each ZSL dataset. Details of hyperparameter search can be found in Ablation Study section.
\subsection{State-Of-The-Art Comparison}
We compare the performance of our model with several SOTA ZSL models in Table \ref{Table:performance}. $A_T$ refers to Traditional ZSL per-class Top-1 Accuracy. $A_U,A_S$ refers to Generalized ZSL unseen and seen class Top-1 per-class Accuracy respectively. Harmonic mean $H=2(A_U*A_S)/(A_U + A_S)$ measures the trade-off between seen and unseen class accuracy.

%\NB{You need a space between name and citation always - you have a large number of inconsistencies on these}
The reported performance of SYNC \cite{SYNC}, ALE \cite{ALE}, DEVISE \cite{DEVISE}, GFZSL \cite{GFZSL} are updated by Xian {\it et al.} \cite{res101}. GDAN \cite{GDAN}, CADA-VAE \cite{CADA-VAE}, TF-VAEGAN \cite{TF-VAEGAN}, LisGAN \cite{LisGAN}, GCM-CF \cite{CFZSL} were updated by GCM-CF \cite{CFZSL}. FREE \cite{FREE} and CNZSL \cite{CNZSL} adopt ``Proposed Split V2.0" already in their paper. Performance of E-PGN \cite{EPGN}, Li {\it et al.} \cite{rethinking} and DVBE \cite{DVBE} on ``Proposed Split V2.0" are fintuned and updated by the author using the published official code of each paper. %\NB{Just as is or did you make any optimization efforts with these?}

Following \cite{CFZSL}, we have not listed models that only report performance on incorrect ``Proposed Split", including f-VAEGAN-D2 \cite{f-VAEGAN-D2}, RELATION NET \cite{RelationNet}, DAZLE \cite{DAZLE}, OCD \cite{OCD}, IZF \cite{IZF}, AGZSL \cite{AGZSL}, IPN \cite{IPN} and CE-GZSL \cite{CE-GZSL}. SOTA models that only report ImageNet performance like DGP \cite{DGP} and HVE \cite{HVE}, or only report transductive ZSL results like SDGN \cite{wu2020self} are also not listed. A detailed discussion can be found in the supplementary material.
%\NB{You need to note - I presume these are what you report in Tab II?}

%Performance of recent SOTA models: GDAN \cite{GDAN}, E-PGN \cite{EPGN}, RFF-GZSL \cite{RFF}, CADA-VAE \cite{CADA-VAE}, Li {\it et al.} \cite{rethinking} are reproduced using the published official code of each paper, detailed information can be found in the supplementary material. RFF-GZSL \cite{RFF} synthesizes unseen class features to train a classifier. Different generated features may have a large influence on the performance. Therefore we further report error bars for the RFF-GZSL model. We notice that CNZSL \cite{CNZSL} published in 2021 have reported their results on "Proposed Split V2.0", therefore we also include their results in Table \ref{Table:performance}.

As can be seen from Table \ref{Table:performance}, our model has reached SOTA performance on the AWA2, AWA1 and APY datasets. Especially on the APY dataset where the dataset has a significant class-imbalance data distribution, our model outperforms SOTA results by a large margin. Our model has a somewhat lower performance on the CUB and SUN datasets. This may be due to the fact that CUB and SUN are fine-grained datasets and our latent embedding network cannot efficiently capture small differences between classes in these datasets.

\subsection{Training Speed Comparison}
The average training times of SOTA models on each dataset are reported in Table \ref{Table:time}. With the help of our adjusted triplet loss, our model can be trained within as little as only a few minutes on all ZSL datasets. The only model that trains faster than ours is CNZSL \cite{CNZSL}, however, our model achieves better performance than theirs on all ZSL datasets except SUN. GDAN, E-PGN and CADA-VAE have similar training times as our model but with a lower performance on GZSL task. %RFF-GZSL is a generative model that requires a much longer training time to achieve better performance. 
%\NB{on all the standard ZSL datasets compared in this study?.} \NB{Given the time for E-PGN is very close, you should probably say something about its performance specifically (seems good on ZSL, but behind on GZSL - the more realistic setting?}
\begin{table}[t]
\centering
\footnotesize
\setlength\tabcolsep{2pt}
\begin{tabular}{c||c|c|c|c|c} \hline
Dataset & CUB & SUN & AWA2 & AWA1 & APY \\\hline
GDAN \cite{GDAN} & 8min & 18min & 14min & - & 7min\\
CADA-VAE \cite{CADA-VAE} & 3min & 5min & 6min & 6min & -\\
E-PGN \cite{EPGN}& 5min & - & 9min & 8min & - \\ 
DVBE \cite{DVBE} & 180min & - & 540min & - & 210min\\ 
%RFF-GZSL \cite{RFF}5-nn & 180min & 180min & - & 210min & - \\ 
CNZSL \cite{CNZSL} & 0.5min & 0.5min & 0.5min & 0.5min & - \\ \hline
$\mathcal{L}_{BT}$ + GP(\textbf{ours}) & 5min & 8min & 3min & 3min & 2min\\ 
\hline
\end{tabular}
\caption{Average Training Time (minutes) for different models with an NVIDIA RTX 2080 Ti GPU card on each ZSL dataset. The training time of our model is competitive with other models}
\setlength\tabcolsep{6pt}
\label{Table:time}
\end{table}
\begin{table*}[ht!]
\centering
\scriptsize
%\tiny
\setlength\tabcolsep{3pt}
\begin{tabular}{c||c|c c c||c|c c c||c|c c c||c|c c c ||c|c c c}
 \multicolumn{21}{c}{ } \\ \hline
 \multirow{3}{*}{Datasets} & \multicolumn{4}{c||}{CUB} & \multicolumn{4}{c||}{SUN} & \multicolumn{4}{c||}{AWA2} & \multicolumn{4}{c||}{AWA1} & \multicolumn{4}{c}{APY} \\ \cline{2-21}
    & ZSL & \multicolumn{3}{|c||}{GZSL} & ZSL & \multicolumn{3}{|c||}{GZSL} & ZSL & \multicolumn{3}{|c||}{GZSL} & ZSL & \multicolumn{3}{|c||}{GZSL} & ZSL & \multicolumn{3}{|c}{GZSL}\\ \cline{2-21}
    & $A_T$ & $A_{U}$ & $A_{S} $ & $H$ & $A_T$ & $A_{U}$ & $A_{S}$ & $H$ & $A_T$ & $A_{U}$ & $A_{S}$ & $H$ & $A_T$ & $A_{U}$ & $A_{S}$ & $H$ & $A_T$ & $A_{U}$ & $A_{S}$ & $H$\\ \hline
KRR & 20.8 & 14.6 & 24.1 & 18.8 & 40.0 & 29.8 & 19.3 & 23.4 & 43.9 & 28.9 & 61.2 & 39.3 & 43.3 & 28.6 & 62.8 & 39.3 & 34.7 & 26.4 & 70.1 & 38.4 \\
$\mathcal{L}_{BT}$ + KRR & 22.1 & 16.4 & 25.5 & 20.0 & 40.1 & 28.5 & 23.0 & 25.5 & 44.2 & 32.9 & 54.4 & 41.0 & 43.7 & 34.8 & 55.6 & 42.8 & 35.3 & 31.3 & 63.9 & 42.0 \\
GP & 53.3 & 42.4 & 46.3 & 44.2 & 61.9 & 51.7 & 32.9 & 40.2 & \textbf{69.2} & 54.7 & \textbf{78.0} & 64.3 & 69.7 & 57.6 & \textbf{73.9} & 64.7 & 38.3 & 31.5 & 72.6 & 44.0\\
$\mathcal{L}_{T} + $ GP& 57.0 & 48.1 & 51.3 & 49.7 & 57.4 & 48.3 & 25.7 & 33.5 & 64.5 & 56.7 & 75.8 & 64.9 & 66.6 & 59.7 & 73.3 & 65.8 & 40.5 & 34.1 & \textbf{75.2} & 47.0 \\
$\mathcal{L}_{BT}$ + GP(\textbf{ours}) & \textbf{59.9} & \textbf{50.1} & \textbf{56.3} & \textbf{53.1} & \textbf{63.2} & \textbf{50.4} & \textbf{34.8} & \textbf{41.2} & 68.6 & \textbf{62.2} & 76.7 & \textbf{68.7} & \textbf{70.1} & \textbf{64.5} & 73.3 & \textbf{68.6} & \textbf{47.1} & \textbf{42.8} & 64.3 & \textbf{51.4} \\ 
\hline
\end{tabular}
\caption{Ablation Study on different model structures. Our proposed $\mathcal{L}_{BT}$ + GP model performs consistently better than Kernel Ridge Regression (KRR) models, the Gaussian Process (GP) model and the GP model with the original triplet loss $\mathcal{L}_{T}$ + GP.}
\setlength\tabcolsep{6pt}
\label{Table:perf_ablation}
\end{table*}

\begin{table*}[ht!]
\centering
\scriptsize
%\tiny
\setlength\tabcolsep{3pt}
\begin{tabular}{c|c c c c c c c c c c c c} \hline
class & cow & horse & motorbike & person & pottedplant & sheep & train & tvmonitor & donkey & goat & jetski & statue \\ \hline
Sample Frequency & 2.48\% & 3.81\% & 3.75\% & 63.99\% & 5.50\% & 2.95\% & 2.22\% & 3.77\%  & 1.75\% & 2.06\% & 5.04\% & 2.61\% \\ \hline
$\mathcal{L}_T$ + GP & 7.1 & \textbf{36.9} & 75.7 & 7.4 & 31.4 & 17.9 & 80.1 & 75.9 & 12.2 & \textbf{63.2} & \textbf{51.4} & 13.0 \\
$\mathcal{L}_{BT}$ + GP\textbf{(ours)} & \textbf{15.7} & 35.9 & \textbf{75.8} & \textbf{12.3} &  \textbf{33.9} & \textbf{19.2} & \textbf{84.6} & \textbf{82.3} & \textbf{36.7} & 48.5 & 40.6 & \textbf{28.5} \\
\hline
\end{tabular}
\caption{Ablation Study of per-class accuracy for unseen classes on class-imbalanced dataset APY. Our proposed $\mathcal{L}_{BT}$ + GP model performs better than the GP model with the original triplet loss $\mathcal{L}_{T}$ + GP consistently on most of the classes.}
\setlength\tabcolsep{6pt}
\label{Table:percls_ablation}
\end{table*}
\subsection{Area Under Seen and Unseen Curve (AUSUC)}
For the GZSL problem, models usually have to balance the trade-off between seen and unseen class accuracies, which is measured by the Harmonic mean $H$. Similar to our model, many SOTA models like \cite{CNZSL,CFZSL} introduced a calibration parameter $\gamma$ to account for the trade-off. Recently, Yue {\it et al.} \cite{CFZSL} proposed to utilize $\gamma$ and plot Area Under Seen and Unseen Curve (AUSUC). Such a figure can provide a more detailed measure of the seen and unseen class trade-off. We compare AUSUC curve of our model with CADA-VAE \cite{CADA-VAE} and CNZSL \cite{CNZSL}, which have official code available.

%\begin{figure}
%	\centering
%	\begin{minipage}{1.7in}
%		\includegraphics[width=1.7in]{./images/AWA2-SUAUC.pdf}
%	\end{minipage}
%	\begin{minipage}{1.7in}
%		\includegraphics[width=1.7in]{./images/AWA1-SUAUC.pdf}
%	\end{minipage}
%		\begin{minipage}{1.7in}
%		\includegraphics[width=1.7in]{./images/APY-SUAUC.pdf}
%	\end{minipage}
%	%\begin{minipage}{1.7in}
%	%	\includegraphics[width=1.7in]{./images/SUN-SUAUC.pdf}
%	%\end{minipage}
%	%\begin{minipage}{1.7in}
%	%	\includegraphics[width=1.7in]{./images/CUB-SUAUC.pdf}
%	%\end{minipage}
%	\caption{Comparison of Area Under Seen and Unseen Curve (AUSUC) for each dataset with CNZSL \cite{CNZSL} and CADA-VAE \cite{CADA-VAE}. Our model is performing consistently better than CNZSL and CADA-VAE on class-imbalanced dataset AWA2, AWA1 and APY.}
%	\label{Fig:AUSUC}
%\end{figure}

\begin{figure}
    \centering
    \includegraphics[width=3.2in]{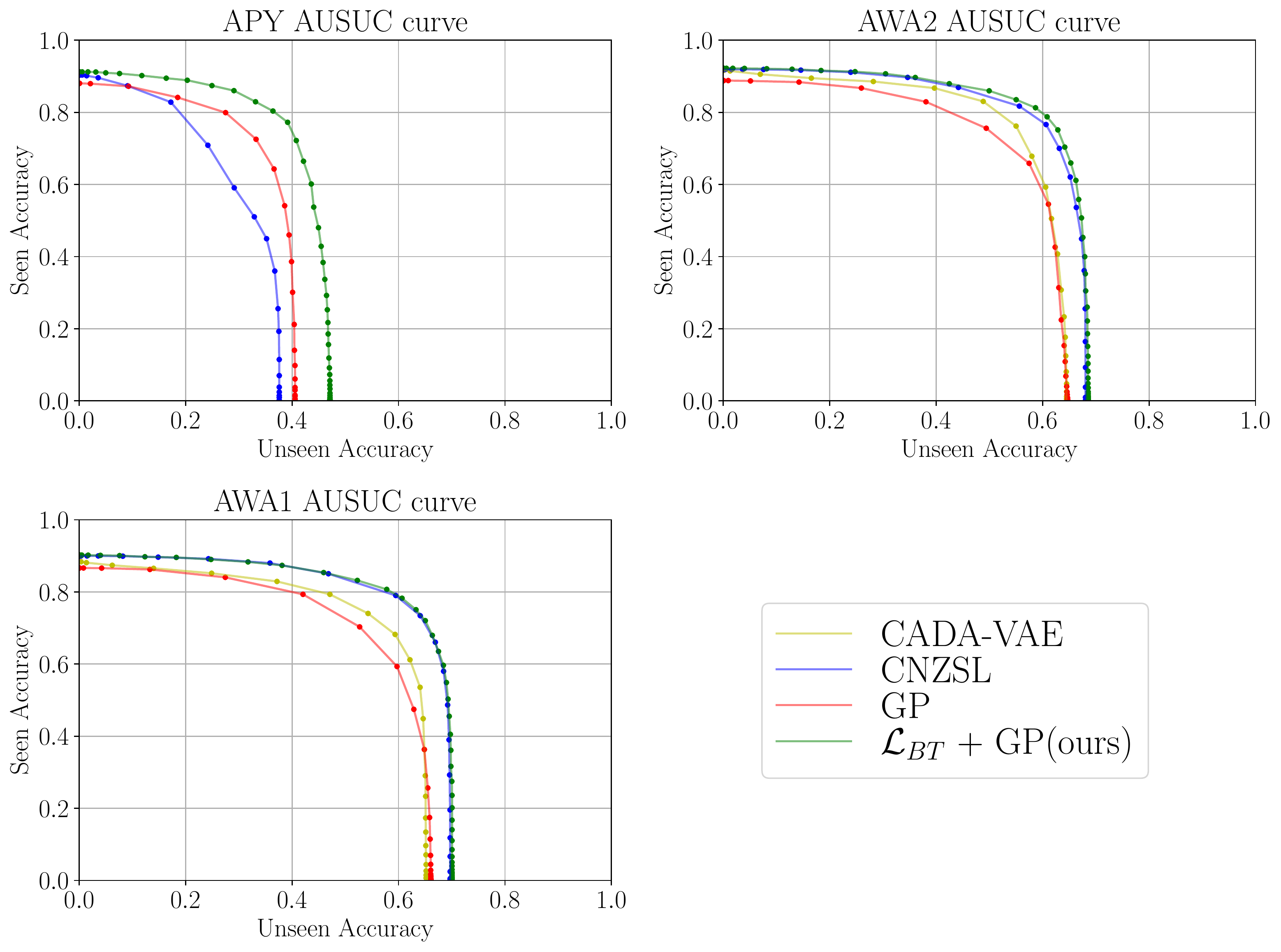}
    \caption{Comparison of Area Under Seen and Unseen Curve (AUSUC) for each dataset with CNZSL \cite{CNZSL} and CADA-VAE \cite{CADA-VAE}. Our model is performing consistently better than CNZSL and CADA-VAE on class-imbalanced datasets AWA2, AWA1 and APY.}
    \label{Fig:AUSUC}
\end{figure}

From Figure \ref{Fig:AUSUC} we can see that our model performs consistently better than CNZSL and CADA-VAE on class-imbalanced datasets AWA2 and APY, and competitive with CNZSL on AWA1 dataset. %\NB{awa1 better than or equal to? seems very close at times, btu not worse - or ok as is?}

\subsection{Ablation Study}
\textbf{Model Structure Ablation:} We compare our model's performance with several similar models' structures in Table \ref{Table:perf_ablation}. We report the performance of the Kernel Ridge Regression (KRR) model on feature space (as a baseline for GPR), KRR on latent space which is trained with our proposed $\mathcal{L}_{BT}$ loss, the Gaussian Process (GP) model on feature space and the GP model on latent space trained with the original triplet loss $\mathcal{L}_{T}$. Our model performs consistently better than all the other baseline approaches on each ZSL dataset.

\textbf{Hyperparameter Ablation:} We analyze the influence of two main hyperparameters on the performance of our model. These hyperparameters are the clip value that is used for preprocessing feature vectors and the threshold $\Delta$ used in triplet loss. 
%Feature vectors that we use are clipped by a fixed value and normalized to $[0,1]$, we analyze the influence of different clipping values on our model's performance. $\Delta$ is another hyperparameter in the triplet loss, which serves as a threshold that balances the intra-class sample distance and the inter-class sample distance. We report the influence of these two hyperparameters on the performance of our model in Figure \ref{Fig:Ablation}. Further comparisons can be found in the supplementary material.
%\NB{You need to ablate against going to standard triplet loss rather than your zero shot class imbalanced version, and other variants around it as this is a key contribution.}
\begin{figure}
	\centering
	\begin{minipage}{1.6in}
		\includegraphics[width=1.6in]{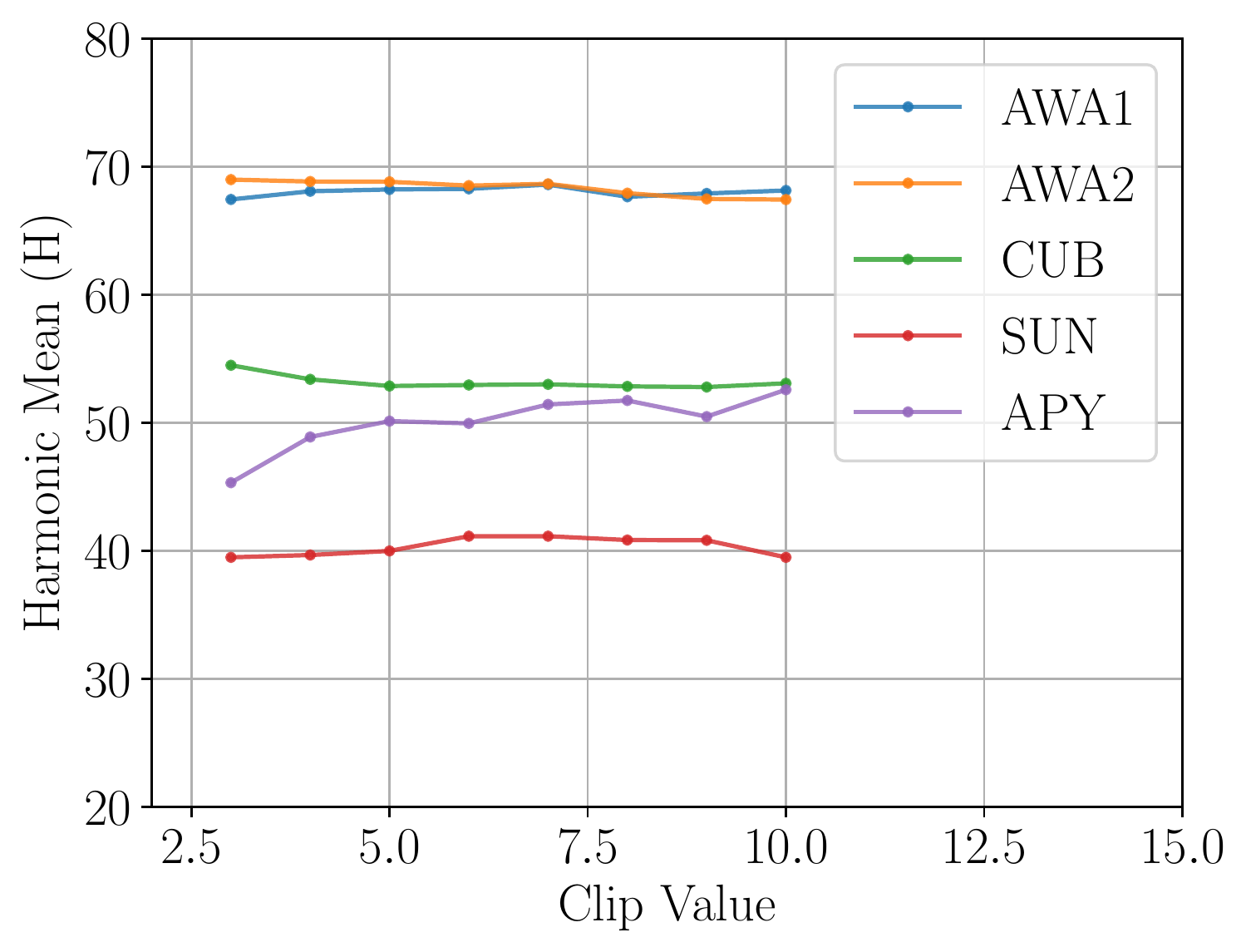}
	\end{minipage}
	\begin{minipage}{1.6in}
		\includegraphics[width=1.6in]{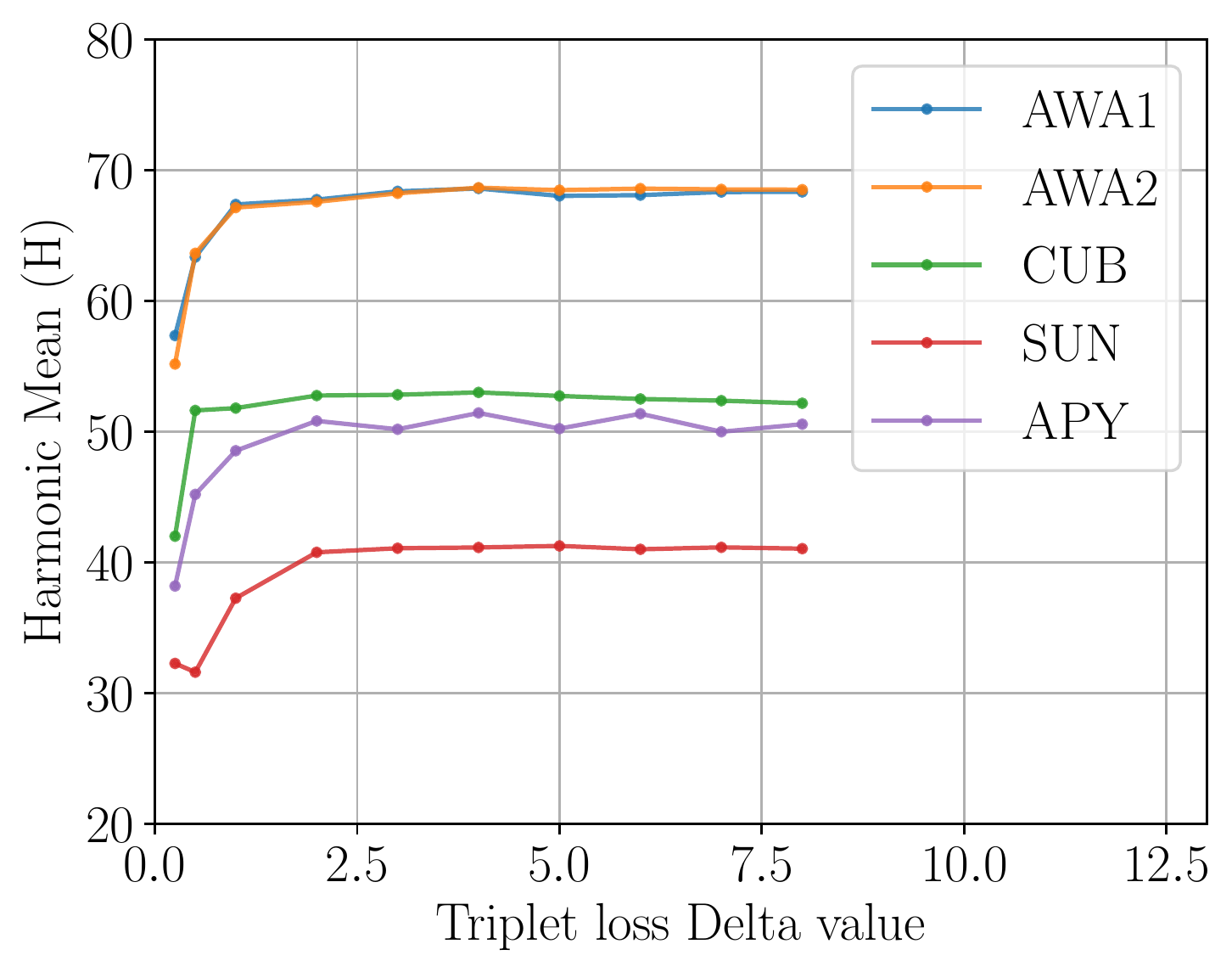}
	\end{minipage}
	\caption{Harmonic mean $H$ for each dataset influenced by the hyperparameter Clip value in preprocessing (left) and $\Delta$ in triplet loss(right). Our model is stable w.r.t these hyperparameters}
	\label{Fig:Ablation}
\end{figure}
As can be seen from Figure \ref{Fig:Ablation}, our model is not sensitive to the clip number that is used in data preprocessing when the clip number is in the range $[4,9]$. As long as $\Delta>3$, the performance of our model on each dataset is relatively stable. %$H$ on the APY dataset seems to be unstable with respect to $\Delta$, but only within a range of $\pm1.0$.

\textbf{Per-Class Accuracy Ablation:} In Table \ref{Table:percls_ablation}, we also provide per-class accuracy of our model $\mathcal{L}_{BT}+GP$ and traditional triplet loss $\mathcal{L}_T$ + GP, on unseen classes on the APY dataset. As can be seen from the table, the APY dataset has a highly class-imbalanced unseen class distribution, where 64\% of the unseen test samples come from class ``person". Our model performs consistently better than traditional triplet loss models on most of the classes.

\section{Conclusion}
In this work, we propose a novel model that combines a Neural Network and Gaussian Process regression to tackle the problems of ZSL and Generalized ZSL. We propose a NN model that projects feature vectors into a latent embedding space and generates latent prototypes of seen classes. A GP model is then trained to predict prototypes of unseen classes. Finally, a ZSL classifier is constructed using the prototypes.

We trained our NN model with a Class-Balanced Triplet loss that mitigates the problem of class imbalance in ZSL datasets. Experiments demonstrate that our model, though employing a simple design, can reach SOTA performance on the class-imbalanced ZSL datasets AWA1, AWA2 and APY in the Generalized ZSL setting.
\clearpage
\bibliographystyle{./IEEEtran}
\bibliography{./ref.bib}

\begin{thebibliography}{10}
\providecommand{\url}[1]{#1}
\csname url@samestyle\endcsname
\providecommand{\newblock}{\relax}
\providecommand{\bibinfo}[2]{#2}
\providecommand{\BIBentrySTDinterwordspacing}{\spaceskip=0pt\relax}
\providecommand{\BIBentryALTinterwordstretchfactor}{4}
\providecommand{\BIBentryALTinterwordspacing}{\spaceskip=\fontdimen2\font plus
\BIBentryALTinterwordstretchfactor\fontdimen3\font minus
  \fontdimen4\font\relax}
\providecommand{\BIBforeignlanguage}[2]{{%
\expandafter\ifx\csname l@#1\endcsname\relax
\typeout{** WARNING: IEEEtran.bst: No hyphenation pattern has been}%
\typeout{** loaded for the language `#1'. Using the pattern for}%
\typeout{** the default language instead.}%
\else
\language=\csname l@#1\endcsname
\fi
#2}}
\providecommand{\BIBdecl}{\relax}
\BIBdecl

\bibitem{conse}
M.~{Norouzi}, T.~{Mikolov}, S.~{Bengio}, Y.~{Singer}, J.~{Shlens}, A.~{Frome},
  G.~S. {Corrado}, and J.~{Dean}, ``Zero-shot learning by convex combination of
  semantic embeddings,'' \emph{arXiv preprint arXiv:1312.5650}, 2013.

\bibitem{SYNC}
S.~Changpinyo, W.-L. Chao, B.~Gong, and F.~Sha, ``Synthesized classifiers for
  zero-shot learning,'' in \emph{Proceedings of the IEEE conference on computer
  vision and pattern recognition}, 2016, pp. 5327--5336.

\bibitem{GFZSL}
V.~K. Verma and P.~Rai, ``A simple exponential family framework for zero-shot
  learning,'' in \emph{Machine Learning and Knowledge Discovery in Databases},
  M.~Ceci, J.~Hollm{\'e}n, L.~Todorovski, C.~Vens, and S.~D{\v{z}}eroski,
  Eds.\hskip 1em plus 0.5em minus 0.4em\relax Cham: Springer International
  Publishing, 2017, pp. 792--808.

\bibitem{GDAN}
H.~Huang, C.~Wang, P.~S. Yu, and C.-D. Wang, ``Generative dual adversarial
  network for generalized zero-shot learning,'' in \emph{The IEEE Conference on
  Computer Vision and Pattern Recognition (CVPR)}, June 2019.

\bibitem{f-CLSWGAN}
Y.~{Xian}, T.~{Lorenz}, B.~{Schiele}, and Z.~{Akata}, ``Feature generating
  networks for zero-shot learning,'' in \emph{2018 IEEE/CVF Conference on
  Computer Vision and Pattern Recognition}, 2018, pp. 5542--5551.

\bibitem{GAN}
I.~J. Goodfellow, J.~Pouget-Abadie, M.~Mirza, B.~Xu, D.~Warde-Farley, S.~Ozair,
  A.~Courville, and Y.~Bengio, ``Generative adversarial networks,'' 2014.

\bibitem{VAE}
D.~P. {Kingma} and M.~{Welling}, ``Auto-encoding variational bayes,''
  \emph{arXiv preprint arXiv:1312.6114}, 2013.

\bibitem{tripletloss}
Y.~Le~Cacheux, H.~Le~Borgne, and M.~Crucianu, ``Modeling inter and intra-class
  relations in the triplet loss for zero-shot learning,'' in \emph{the IEEE
  International Conference on Computer Vision (ICCV)}, ser. ICCV, October 2019.

\bibitem{RFF}
Z.~Han, Z.~Fu, and J.~Yang, ``Learning the redundancy-free features for
  generalized zero-shot object recognition,'' in \emph{IEEE/CVF Conference on
  Computer Vision and Pattern Recognition (CVPR)}, June 2020.

\bibitem{FREE}
S.~Chen, W.~Wang, B.~Xia, Q.~Peng, X.~You, F.~Zheng, and L.~Shao, ``Free:
  Feature refinement for generalized zero-shot learning,'' in \emph{Proceedings
  of the IEEE/CVF International Conference on Computer Vision}, 2021, pp.
  122--131.

\bibitem{DVBE}
S.~Min, H.~Yao, H.~Xie, C.~Wang, Z.~Zha, and Y.~Zhang, ``Domain-aware visual
  bias eliminating for generalized zero-shot learning,'' \emph{2020 IEEE/CVF
  Conference on Computer Vision and Pattern Recognition (CVPR)}, pp.
  12\,661--12\,670, 2020.

\bibitem{CE-GZSL}
Z.~Han, Z.~Fu, S.~Chen, and J.~Yang, ``Contrastive embedding for generalized
  zero-shot learning,'' in \emph{Proceedings of the IEEE/CVF Conference on
  Computer Vision and Pattern Recognition (CVPR)}, June 2021, pp. 2371--2381.

\bibitem{openlongtailrecognition}
Z.~Liu, Z.~Miao, X.~Zhan, J.~Wang, B.~Gong, and S.~X. Yu, ``Large-scale
  long-tailed recognition in an open world,'' in \emph{IEEE Conference on
  Computer Vision and Pattern Recognition (CVPR)}, 2019.

\bibitem{buda2018systematic}
M.~Buda, A.~Maki, and M.~A. Mazurowski, ``A systematic study of the class
  imbalance problem in convolutional neural networks,'' \emph{Neural Networks},
  vol. 106, pp. 249--259, 2018.

\bibitem{cui2019classbalancedloss}
Y.~Cui, M.~Jia, T.-Y. Lin, Y.~Song, and S.~Belongie, ``Class-balanced loss
  based on effective number of samples,'' in \emph{CVPR}, 2019.

\bibitem{APY}
A.~{Farhadi}, I.~{Endres}, D.~{Hoiem}, and D.~{Forsyth}, ``Describing objects
  by their attributes,'' in \emph{2009 IEEE Conference on Computer Vision and
  Pattern Recognition}, 6 2009, pp. 1778--1785.

\bibitem{res101}
Y.~{Xian}, C.~H. {Lampert}, B.~{Schiele}, and Z.~{Akata}, ``Zero-shot
  learning—a comprehensive evaluation of the good, the bad and the ugly,''
  \emph{IEEE Transactions on Pattern Analysis and Machine Intelligence},
  vol.~41, no.~9, pp. 2251--2265, Sep. 2019.

\bibitem{gpbook}
\BIBentryALTinterwordspacing
C.~Rasmussen, C.~Williams, M.~Press, F.~Bach, and P.~(Firm), \emph{Gaussian
  Processes for Machine Learning}, ser. Adaptive computation and machine
  learning.\hskip 1em plus 0.5em minus 0.4em\relax MIT Press, 2006. [Online].
  Available: \url{https://books.google.com.au/books?id=GhoSngEACAAJ}
\BIBentrySTDinterwordspacing

\bibitem{PAC2012}
T.~Suzuki, ``Pac-bayesian bound for gaussian process regression and multiple
  kernel additive model,'' \emph{J. Mach. Learn. Res. Wrkshp Conf. Proc.},
  vol.~23, 01 2012.

\bibitem{ALE}
Z.~Akata, F.~Perronnin, Z.~Harchaoui, and C.~Schmid, ``Label-embedding for
  image classification,'' \emph{IEEE transactions on pattern analysis and
  machine intelligence}, vol.~38, no.~7, pp. 1425--1438, 2015.

\bibitem{Snell2017PrototypicalNF}
J.~Snell, K.~Swersky, and R.~S. Zemel, ``Prototypical networks for few-shot
  learning,'' in \emph{NIPS}, 2017.

\bibitem{prototype1}
R.~Boney and A.~Ilin, ``Semi-supervised few-shot learning with prototypical
  networks,'' \emph{CoRR abs/1711.10856}, 2017.

\bibitem{prototype2}
T.~Gao, X.~Han, Z.~Liu, and M.~Sun, ``Hybrid attention-based prototypical
  networks for noisy few-shot relation classification,'' in \emph{Proceedings
  of the AAAI Conference on Artificial Intelligence}, vol.~33, 2019, pp.
  6407--6414.

\bibitem{transductive0}
O.~Chapelle, B.~Scholkopf, and E.~A.~Zien, ``Semi-supervised learning
  (chapelle, o. et al., eds.; 2006) [book reviews],'' vol.~20, no.~3, pp.
  542--542, 2009.

\bibitem{rethinking}
K.~Li, M.~R. Min, and Y.~Fu, ``Rethinking zero-shot learning: A conditional
  visual classification perspective,'' in \emph{Proceedings of the IEEE
  International Conference on Computer Vision}, 2019, pp. 3583--3592.

\bibitem{transductive1}
Y.~Meng and Y.~Guo, ``Zero-shot classification with discriminative semantic
  representation learning,'' in \emph{2017 IEEE Conference on Computer Vision
  and Pattern Recognition (CVPR)}, 2017.

\bibitem{centerloss}
Y.~Wen, K.~Zhang, Z.~Li, and Y.~Qiao, ``A discriminative feature learning
  approach for deep face recognition,'' in \emph{European conference on
  computer vision}.\hskip 1em plus 0.5em minus 0.4em\relax Springer, 2016, pp.
  499--515.

\bibitem{GPRZSL2}
Y.~Dolma and V.~P. Namboodiri, ``Using gaussian processes to improve zero-shot
  learning with relative attributes,'' in \emph{Computer Vision -- ACCV 2016},
  S.-H. Lai, V.~Lepetit, K.~Nishino, and Y.~Sato, Eds.\hskip 1em plus 0.5em
  minus 0.4em\relax Cham: Springer International Publishing, 2017, pp.
  150--164.

\bibitem{GPRZSL3}
\BIBentryALTinterwordspacing
T.~Mukherjee and T.~Hospedales, ``{Gaussian Visual-Linguistic Embedding for
  Zero-Shot Recognition},'' in \emph{Proceedings of the 2016 Conference on
  Empirical Methods in Natural Language Processing}.\hskip 1em plus 0.5em minus
  0.4em\relax Austin, Texas: Association for Computational Linguistics, Nov.
  2016, pp. 912--918. [Online]. Available:
  \url{https://www.aclweb.org/anthology/D16-1089}
\BIBentrySTDinterwordspacing

\bibitem{GPRZSL}
M.~Elhoseiny, B.~Saleh, and A.~Elgammal, ``Write a classifier: Zero-shot
  learning using purely textual descriptions,'' in \emph{Proceedings of the
  IEEE International Conference on Computer Vision}, 2013, pp. 2584--2591.

\bibitem{inbalancetriplet}
C.~Huang, Y.~Li, C.~C. Loy, and X.~Tang, ``Deep imbalanced learning for face
  recognition and attribute prediction,'' \emph{IEEE transactions on pattern
  analysis and machine intelligence}, vol.~42, no.~11, pp. 2781--2794, 2019.

\bibitem{tripletbad1}
K.~Sohn, ``Improved deep metric learning with multi-class n-pair loss
  objective,'' in \emph{Advances in Neural Information Processing Systems},
  D.~Lee, M.~Sugiyama, U.~Luxburg, I.~Guyon, and R.~Garnett, Eds.,
  vol.~29.\hskip 1em plus 0.5em minus 0.4em\relax Curran Associates, Inc.,
  2016.

\bibitem{calibration}
Y.~Le~Cacheux, H.~Le~Borgne, and M.~Crucianu, ``From classical to generalized
  zero-shot learning: A simple adaptation process,'' in \emph{International
  Conference on Multimedia Modeling}.\hskip 1em plus 0.5em minus 0.4em\relax
  Springer, 2019, pp. 465--477.

\bibitem{CNZSL}
I.~Skorokhodov and M.~Elhoseiny, ``Class normalization for (continual)?
  generalized zero-shot learning,'' in \emph{Proceedings of the International
  Conference on Learning Representations (ICLR)}, 2021.

\bibitem{CUB}
C.~Wah, S.~Branson, P.~Welinder, P.~Perona, and S.~Belongie, ``{The
  Caltech-UCSD Birds-200-2011 Dataset},'' California Institute of Technology,
  Tech. Rep. CNS-TR-2011-001, 2011.

\bibitem{SUN}
G.~Patterson and J.~Hays, ``Sun attribute database: Discovering, annotating,
  and recognizing scene attributes,'' in \emph{2012 IEEE Conference on Computer
  Vision and Pattern Recognition}.\hskip 1em plus 0.5em minus 0.4em\relax IEEE,
  2012, pp. 2751--2758.

\bibitem{DEVISE}
A.~Frome, G.~S. Corrado, J.~Shlens, S.~Bengio, J.~Dean, M.~Ranzato, and
  T.~Mikolov, ``Devise: A deep visual-semantic embedding model,'' in
  \emph{Advances in neural information processing systems}, 2013, pp.
  2121--2129.

\bibitem{CFZSL}
\BIBentryALTinterwordspacing
Z.~Yue, T.~Wang, H.~Zhang, Q.~Sun, and X.~Hua, ``Counterfactual zero-shot and
  open-set visual recognition,'' \emph{CoRR}, vol. abs/2103.00887, 2021.
  [Online]. Available: \url{https://arxiv.org/abs/2103.00887}
\BIBentrySTDinterwordspacing

\bibitem{CADA-VAE}
E.~{Schönfeld}, S.~{Ebrahimi}, S.~{Sinha}, T.~{Darrell}, and Z.~{Akata},
  ``Generalized zero- and few-shot learning via aligned variational
  autoencoders,'' in \emph{2019 IEEE/CVF Conference on Computer Vision and
  Pattern Recognition (CVPR)}, 2019, pp. 8239--8247.

\bibitem{TF-VAEGAN}
S.~Narayan, A.~Gupta, F.~S. Khan, C.~G. Snoek, and L.~Shao, ``Latent embedding
  feedback and discriminative features for zero-shot classification,'' in
  \emph{ECCV}, 2020.

\bibitem{LisGAN}
J.~Li, M.~Jing, K.~Lu, Z.~Ding, L.~Zhu, and Z.~Huang, ``Leveraging the
  invariant side of generative zero-shot learning,'' in \emph{Proceedings of
  the IEEE/CVF Conference on Computer Vision and Pattern Recognition}, 2019,
  pp. 7402--7411.

\bibitem{EPGN}
Y.~Yu, Z.~Ji, J.~Han, and Z.~Zhang, ``Episode-based prototype generating
  network for zero-shot learning,'' in \emph{IEEE/CVF Conference on Computer
  Vision and Pattern Recognition (CVPR)}, June 2020.

\bibitem{gpytorch}
J.~R. {Gardner}, G.~{Pleiss}, D.~{Bindel}, K.~Q. {Weinberger}, and A.~G.
  {Wilson}, ``Gpytorch: Blackbox matrix-matrix gaussian process inference with
  gpu acceleration,'' \emph{arXiv preprint arXiv:1809.11165}, 2018.

\bibitem{Adam}
\BIBentryALTinterwordspacing
D.~P. Kingma and J.~Ba, ``Adam: {A} method for stochastic optimization,'' in
  \emph{3rd International Conference on Learning Representations, {ICLR} 2015,
  San Diego, CA, USA, May 7-9, 2015, Conference Track Proceedings}, Y.~Bengio
  and Y.~LeCun, Eds., 2015. [Online]. Available:
  \url{http://arxiv.org/abs/1412.6980}
\BIBentrySTDinterwordspacing

\bibitem{f-VAEGAN-D2}
Y.~{Xian}, S.~{Sharma}, B.~{Schiele}, and Z.~{Akata}, ``F-vaegan-d2: A feature
  generating framework for any-shot learning,'' in \emph{2019 IEEE/CVF
  Conference on Computer Vision and Pattern Recognition (CVPR)}, 2019, pp.
  10\,267--10\,276.

\bibitem{RelationNet}
F.~{Sung}, Y.~{Yang}, L.~{Zhang}, T.~{Xiang}, P.~H.~S. {Torr}, and T.~M.
  {Hospedales}, ``Learning to compare: Relation network for few-shot
  learning,'' in \emph{2018 IEEE/CVF Conference on Computer Vision and Pattern
  Recognition}, 2018, pp. 1199--1208.

\bibitem{DAZLE}
D.~Huynh and E.~Elhamifar, ``Fine-grained generalized zero-shot learning via
  dense attribute-based attention,'' in \emph{IEEE/CVF Conference on Computer
  Vision and Pattern Recognition (CVPR)}, June 2020.

\bibitem{OCD}
R.~Keshari, R.~Singh, and M.~Vatsa, ``Generalized zero-shot learning via
  over-complete distribution,'' in \emph{IEEE/CVF Conference on Computer Vision
  and Pattern Recognition (CVPR)}, June 2020.

\bibitem{IZF}
Y.~{Shen}, J.~{Qin}, L.~{Huang}, L.~{Liu}, F.~{Zhu}, and L.~{Shao},
  ``Invertible zero-shot recognition flows,'' in \emph{European Conference on
  Computer Vision}, 2020, pp. 614--631.

\bibitem{AGZSL}
Y.-Y. Chou, H.-T. Lin, and T.-L. Liu, ``Adaptive and generative zero-shot
  learning,'' in \emph{International Conference on Learning Representations},
  2020.

\bibitem{IPN}
L.~Liu, T.~Zhou, G.~Long, J.~Jiang, X.~Dong, and C.~Zhang, ``Isometric
  propagation network for generalized zero-shot learning,'' in
  \emph{International Conference on Learning Representations}, 2020.

\bibitem{DGP}
M.~{Kampffmeyer}, Y.~{Chen}, X.~{Liang}, H.~{Wang}, Y.~{Zhang}, and E.~P.
  {Xing}, ``Rethinking knowledge graph propagation for zero-shot learning,'' in
  \emph{2019 IEEE/CVF Conference on Computer Vision and Pattern Recognition
  (CVPR)}, 2019, pp. 11\,487--11\,496.

\bibitem{HVE}
S.~{Liu}, J.~{Chen}, L.~{Pan}, C.-W. {Ngo}, T.-S. {Chua}, and Y.-G. {Jiang},
  ``Hyperbolic visual embedding learning for zero-shot recognition,'' in
  \emph{2020 IEEE/CVF Conference on Computer Vision and Pattern Recognition
  (CVPR)}, 2020, pp. 9273--9281.

\bibitem{wu2020self}
J.~{Wu}, T.~{Zhang}, Z.-J. {Zha}, J.~{Luo}, Y.~{Zhang}, and F.~{Wu},
  ``Self-supervised domain-aware generative network for generalized zero-shot
  learning,'' in \emph{2020 IEEE/CVF Conference on Computer Vision and Pattern
  Recognition (CVPR)}, 2020, pp. 12\,767--12\,776.

\end{thebibliography}
\end{document}

\title{Efficient Gaussian Process Model on Class-Imbalanced Datasets for Generalized Zero-Shot Learning -- Supplementary Materials}
%\title{Inexpensive Gaussian Processes for Class-Imbalanced  \\ Generalized Zero-Shot Learning}

\author{\IEEEauthorblockN{Changkun Ye}
\IEEEauthorblockA{
Australian National University \& Data61 CSIRO \\
Canberra, ACT, Australia\\
Email: changkun.ye@anu.edu.au}
\and
\IEEEauthorblockN{Nick Barnes}
\IEEEauthorblockA{Australian National University\\
Canberra, ACT, Australia\\
Email: nick.barnes@anu.edu.au}
\and
\IEEEauthorblockN{Lars Petersson and Russell Tsuchida}
\IEEEauthorblockA{Data61 CSIRO\\
Canberra, ACT, Australia\\
Email: lars.petersson@data61.csiro.au\\
russell.tsuchida@data61.csiro.au}
}

\maketitle

%-------------------------------------------------------------------------
\section{Detailed information of reproduced SOTA models}
In our paper, we reproduced the performance of several SOTA models using their published code on Proposed Split V2.0. They are E-PGN ~\cite{EPGN}, Li {\it et al.} ~\cite{rethinking} and DVBE~\cite{DVBE}. The detailed information for these published models is available in Table \ref{Table:codelink} below.
\begin{table}[h!]
\centering
\footnotesize
\setlength\tabcolsep{2pt}
\begin{tabular}{c|c |c |c } \hline
Model & Conference &Code Link & Time of Retrieval \\\hline
%GDAN~\cite{GDAN} & CVPR 2019 &\href{https://github.com/stevehuanghe/GDAN}{https://github.com/stevehuanghe/GDAN} & Dec 2020\\
%CADA-VAE~\cite{CADA-VAE} & CVPR 2019 & \href{https://github.com/edgarschnfld/CADA-VAE-PyTorch}{https://github.com/edgarschnfld/CADA-VAE-PyTorch} & Dec 2020\\
Li {\it et al.}~\cite{rethinking} & CVPR 19 & https://github.com/kailigo/cvcZSL & Dec 2020 \\
E-PGN~\cite{EPGN}& CVPR 20 &https://github.com/yunlongyu/EPGN & Dec 2020\\ 
DVBE~\cite{DVBE} & CVPR 20&https://github.com/mboboGO/DVBE & Dec 2020\\ 
%RFF-GZSL~\cite{RFF}5-nn & CVPR 2020 & \href{https://github.com/taoting0722/RRF-GZSL}{https://github.com/taoting0722/RRF-GZSL} & Dec 2020\\ 
%CNZSL~\cite{CNZSL} & ICLR 2021&\href{https://github.com/universome/class-norm}{https://github.com/universome/class-norm} & May 2021\\ \hline
\hline
\end{tabular}
\caption{Official, published, code links and time of code retrieval for each reproduced SOTA model in the main paper}
\setlength\tabcolsep{6pt}
\label{Table:codelink}
\end{table}

We reproduce the results by precisely following the instructions provided by the authors of each model, with the exception that we use a different dataset split  \footnote{https://drive.google.com/file/d/1p9gtkuHCCCyjkyezSarCw-1siCSXUykH/view?usp=sharing}{Proposed Splits V2.0}, updated by Xian {\it et al.}~\cite{res101}. {We fine-tune hyperparameters for "Proposed Split V2.0" by parameter search around values recommended for "Proposed Split" in each official code. We note that some models like E-PGN~\cite{EPGN} are sensitive to random seeds, and difficult to fine-tune. Hence, despite our best efforts performance may be sub-optimal.}

The hyperparameters and corresponding values used to reproduce performance for each model Li {\it et al.}~\cite{rethinking}, EPGN~\cite{EPGN}, DVBE~\cite{DVBE} are listed in Table \ref{tab:rp_li}, \ref{tab:rp_EPGN} and \ref{tab:rp_DVBE} respectively.

\begin{table*}[]
    \centering
    \begin{tabular}{c|c c c c c c c } \hline
         Params & ways & shots & lr & opt\_decay & step\_size & log\_file & model\_file \\ \hline
         AWA1 & 16 & 4 & 1e-5 & 1e-4 & 500 & eps\_lr5\_opt4\_ss500\_w16\_s4 & lr5\_opt4\_ss500\_w16\_s4.pt \\
         \hline
    \end{tabular}
    \caption{Hyperparameters used for reproducing Li {\it et al.}~\cite{rethinking} on AWA1 dataset with Proposed Split V2.0. Name of each hyperparameter matches with the published code.}
    \label{tab:rp_li}
\end{table*}

\begin{table*}[]
    \centering
    \begin{tabular}{c|c c c c c c c c c } \hline
         Params & mid\_dim & hid\_dim & lr & epoch & episode & inner\_loop & batch\_size & dropout & manualSeed \\ \hline
         CUB & 1600 & 1800 & 5e-5 & 15 & 100 & 10 & 32 & True & 4196 \\
         AWA1 & 1200 & 1800 & 5e-5 & 30 & 50 & 100 & 100& True & 4198 \\ 
         AWA2 & 1800 & 1800 & 2e-4 & 30 & 50 & 30 & 64 & True & 4198 \\ 
         \hline
    \end{tabular}
    \caption{Hyperparameters used for reproducing EPGN~\cite{EPGN} on CUB, AWA1 and AWA2 datasets with Proposed Split V2.0. Name of each hyperparameter matches with the published code.}
    \label{tab:rp_EPGN}
\end{table*}

\begin{table*}[]
    \centering
    \begin{tabular}{c|| c c c c c c c c c c } \hline 
         Params & batch\_size & lr1 & lr2 & momentum & epochs & epoch\_decay & sigma & weight\_decay & workers & seed\\ \hline \hline
         CUB & 128 & 0.1 & 0.001 & 0.9 & 90 & 30 & 0.5 & 0.0001 & 3 & 5181\\ 
         AWA2 & 128 & 0.1 & 0.001 & 0.9 & 90 & 30 & 0.5 & 0.0001 & 3 & 142 \\
         APY & 128 & 0.1 & 0.001 & 0.9 & 90 & 30 & 0.5 & 0.0001 & 3 & 119\\
         \hline
    \end{tabular}
    \caption{Hyperparameters used for reproducing DVBE~\cite{DVBE} on CUB, AWA2 and APY datasets with Proposed Split V2.0. Name of each hyperparameter matches with the published code.}
    \label{tab:rp_DVBE}
\end{table*}

\section{Ablation Study on model structure}
We report the Area Under Seen and Unseen Curve (AUSUC) of our model along with some alternative model structures. These models include the Kernel Ridge Regression (KRR) model, the KRR model performed on a latent space that was trained with our proposed $\mathcal{L}_{BT}$ triplet loss, the Gaussian Process (GP) model and a GP model performed on a latent space trained using the original triplet loss $\mathcal{L}_{T}$. 

As can be seen from Figure \ref{Fig:AUSUC}, GP based models consistently perform better than KRR based models. Also, our proposed triplet loss $\mathcal{L}_{BT}$ can generally improve the performance of the KRR model as well as the GP model. Our proposed model has improvements in both the seen accuracy and the unseen accuracy compared with other alternative models.
\iffalse
\begin{figure*}
	\centering
	\begin{minipage}{2.1in}
		\includegraphics[width=2.1in]{./images/AWA2-SUAUC-supp.pdf}
	\end{minipage}
	\begin{minipage}{2.1in}
		\includegraphics[width=2.1in]{./images/AWA1-SUAUC-supp.pdf}
	\end{minipage}
		\begin{minipage}{2.1in}
		\includegraphics[width=2.1in]{./images/APY-SUAUC-supp.pdf}
	\end{minipage}
	\begin{minipage}{2.1in}
		\includegraphics[width=2.1in]{./images/SUN-SUAUC-supp.pdf}
	\end{minipage}
	\begin{minipage}{2.1in}
		\includegraphics[width=2.1in]{./images/CUB-SUAUC-supp.pdf}
	\end{minipage}
	\caption{Area Under Seen and Unseen Curve (AUSUC) for different model structures in our ablation study section in the paper. Our model is consistently better than alternative structures. Moreover, KRR and GP models achieve better performance when the latent space is trained with our proposed $\mathcal{L}_{BT}$ loss compared with the original feature space}
	\label{Fig:AUSUC}
\end{figure*}
\fi
\begin{figure}
	\centering
    \includegraphics[width=3.5in]{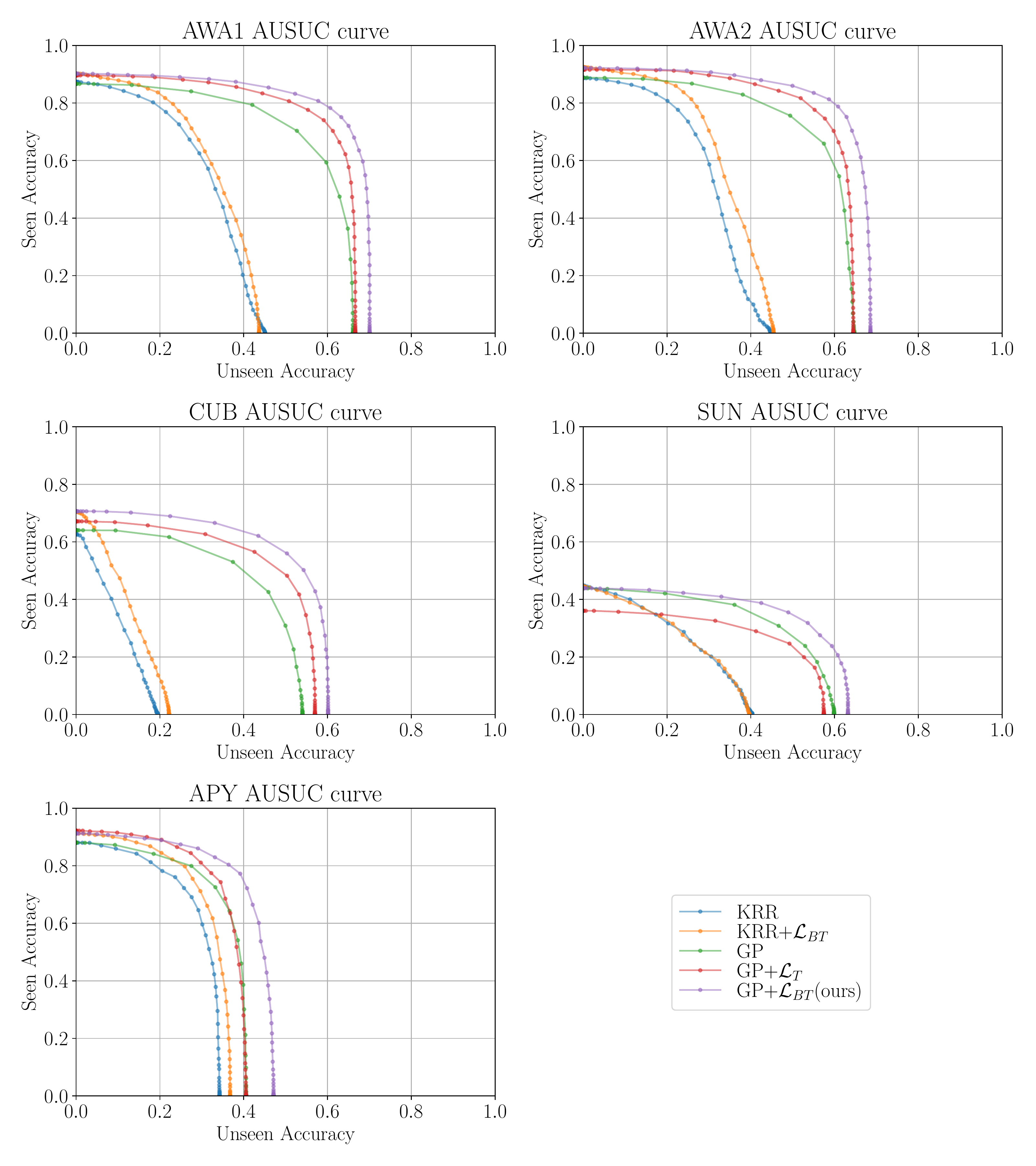}
    \caption{Area Under Seen and Unseen Curve (AUSUC) for different model structures in the ablation study section in the main paper. Our model is consistently better than alternative structures. Moreover, KRR and GP models achieve better performances when the latent space is trained using our proposed $\mathcal{L}_{BT}$ loss compared with the original feature space}
	\label{Fig:AUSUC}
\end{figure}

\begin{table*}[h!]
\centering
\footnotesize
\setlength\tabcolsep{2pt}
\begin{tabular}{c||c|c|c|c|c} \hline
Dataset & CUB & SUN & AWA2 & AWA1 & APY \\\hline
Average feature value & 0.3293 & 0.4413 & 0.4049 & 0.4244 &  0.4459 \\ 
Maximum feature value & 32.95 & 44.83 & 61.00 & 47.21 & 46.55 \\
99.90\% feature value lies in range & $[0.00,6.25]$ & $[0.00,7.81]$ & $[0.00,7.09]$ & $[0.00,7.00]$ & $[0.00,7.74]$ \\
\hline
\end{tabular}
\caption{Analysis of feature vector values. Every dataset has maximal values that are too far away from the average, 99.90\% of the values lies approximately in the range $[0,7]$ for each dataset. Thus, we preprocess feature vectors for each dataset by clipping by 7 and normalize to the range $[0,1]$}
\setlength\tabcolsep{6pt}
\label{Table:feature_dis}
\end{table*}

\section{Normalizing Unbounded Feature Space}
\begin{figure}
	\centering
	\begin{minipage}{1.7in}
		\includegraphics[width=1.7in]{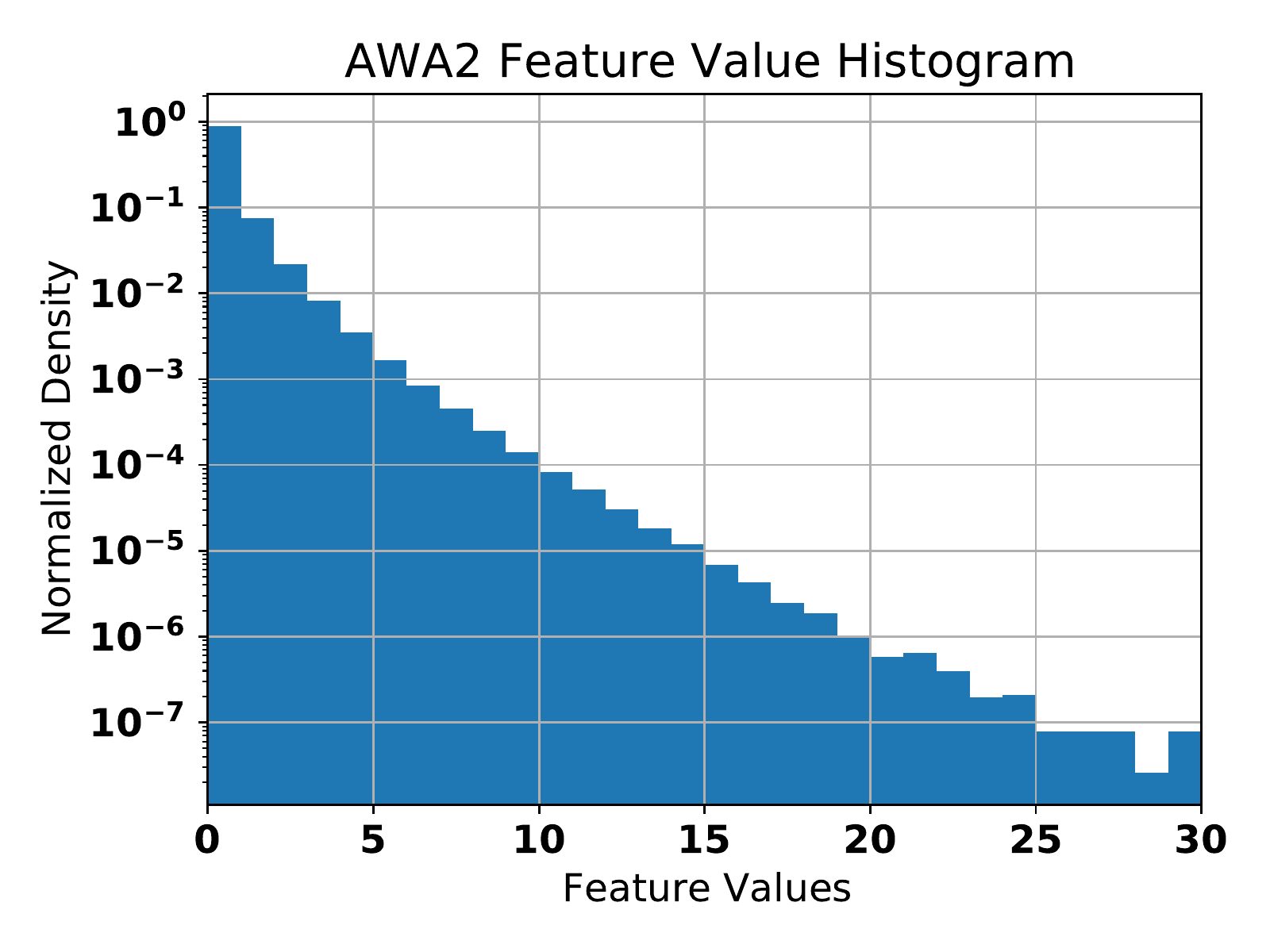}
	\end{minipage}
	\begin{minipage}{1.7in}
		\includegraphics[width=1.7in]{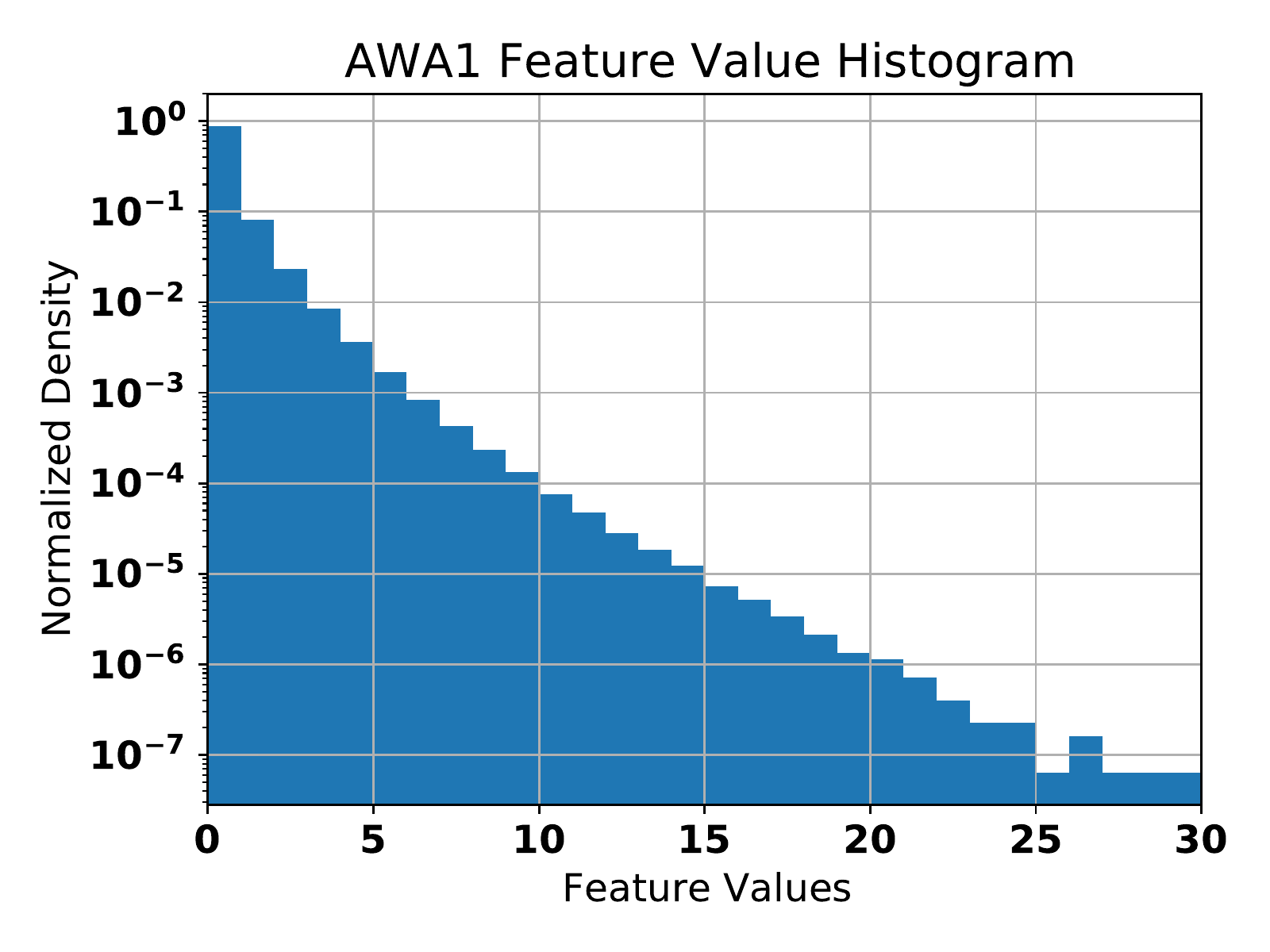}
	\end{minipage}
		\begin{minipage}{1.7in}
		\includegraphics[width=1.7in]{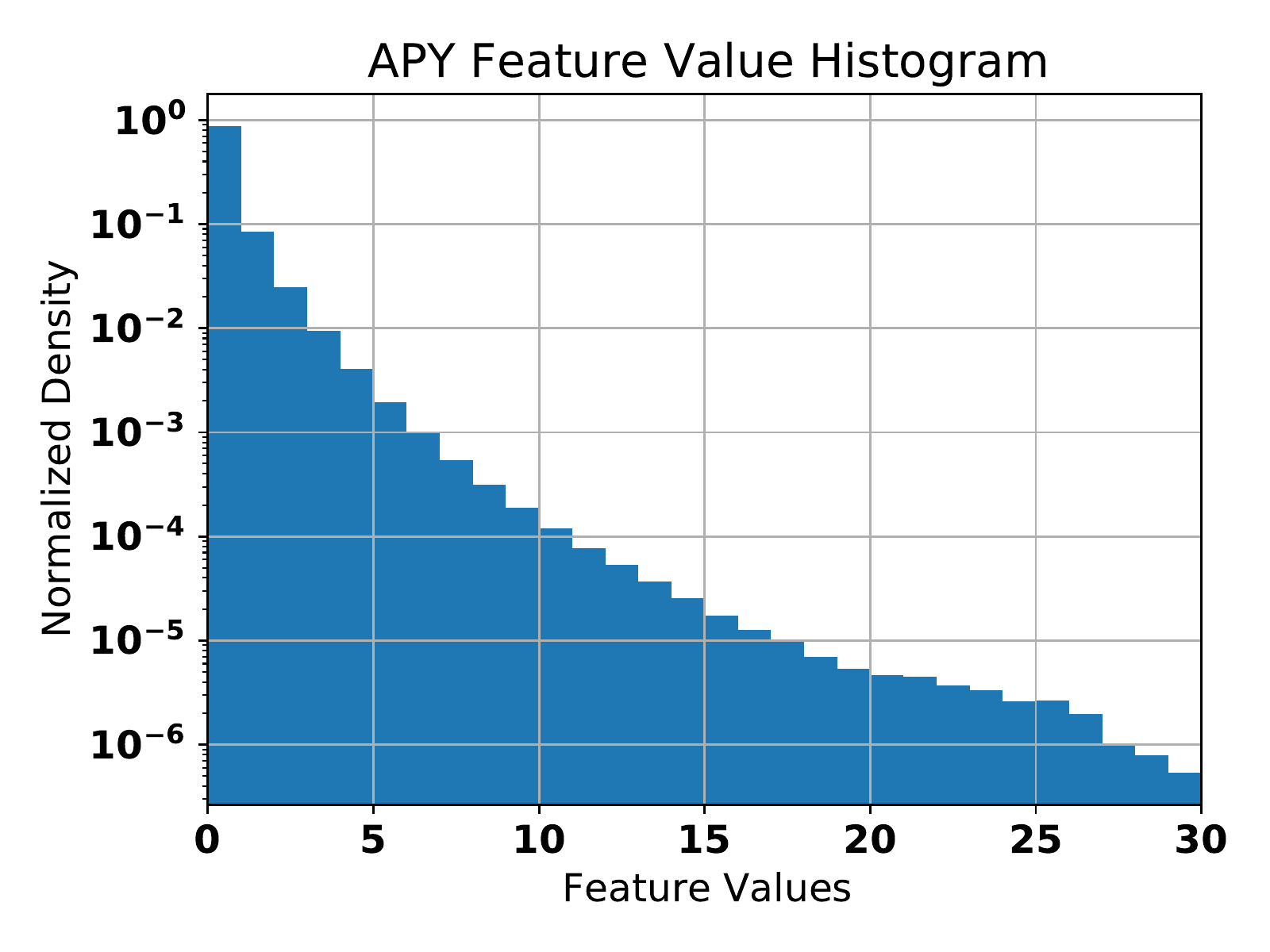}
	\end{minipage}
	\begin{minipage}{1.7in}
		\includegraphics[width=1.7in]{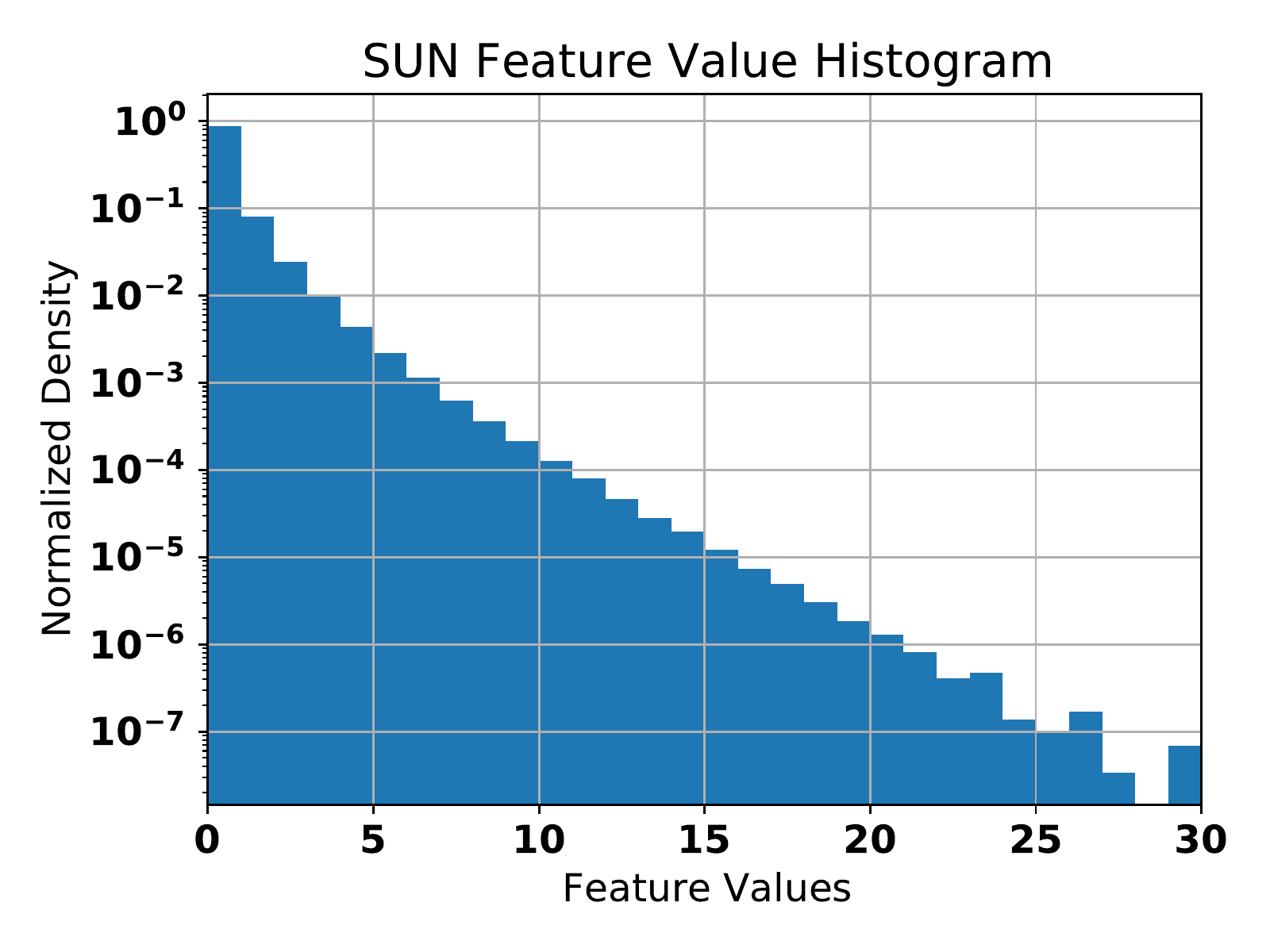}
	\end{minipage}
	\begin{minipage}{1.7in}
		\includegraphics[width=1.7in]{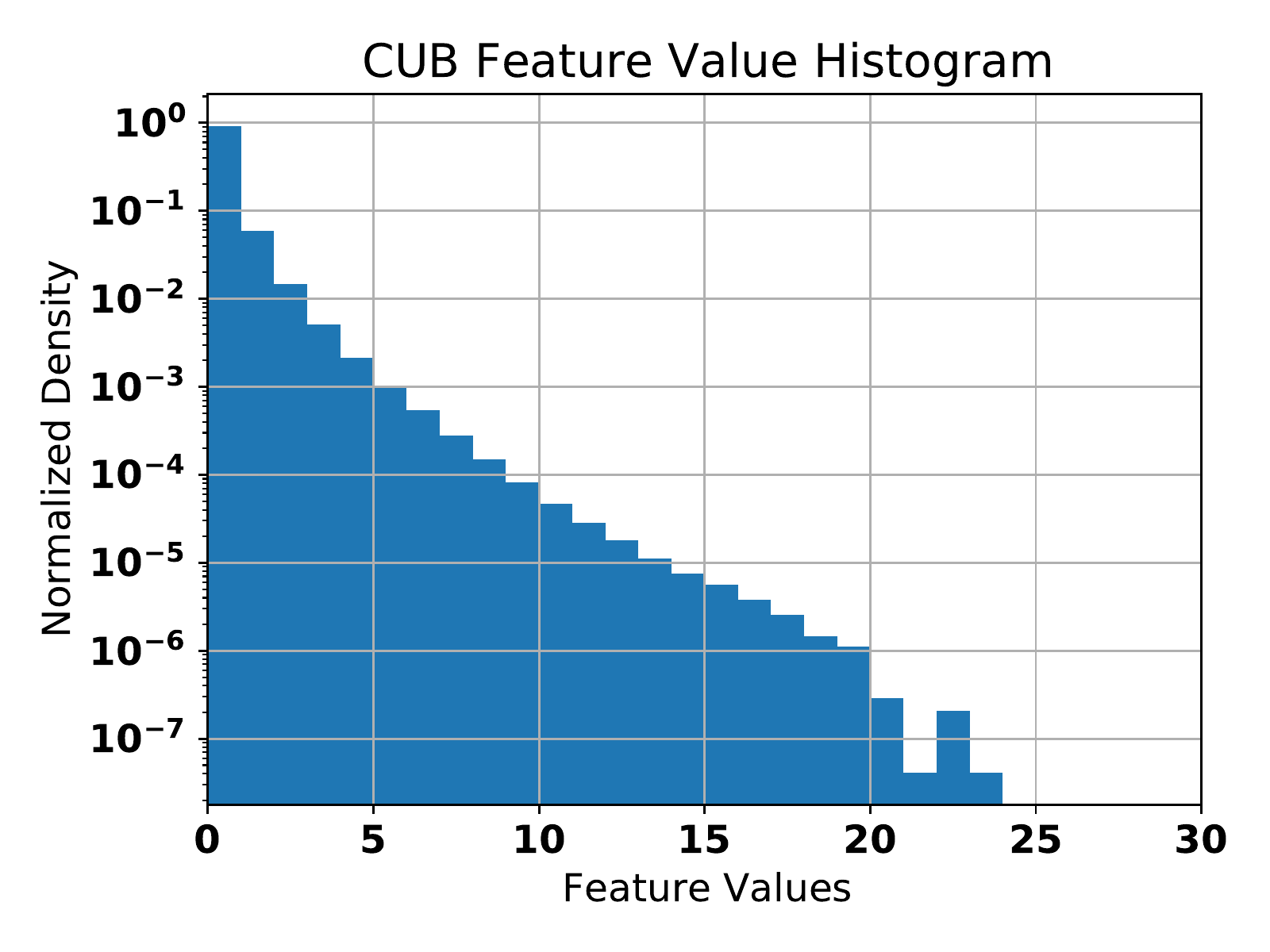}
	\end{minipage}
	\caption{Normalized histogram of feature vector values in each ZSL dataset. The probability density of feature values drop drastically as the feature value increases. Therefore, bounding the feature space by clipping the tail of the density requires modifying only a small amount of the values of the data.} %\NB{I don't like the word about here. You need to be precise or the reviewers may be concerned. Otherwise I like this argument} \russ{Tips for improving plots: (1) Smaller bin width (2) smaller x range. What does ``Density" mean? Are you normalising the histogram so that it is a PDF? Would a logarithmic vertical scale make sense?}
	\label{Fig:hist}
\end{figure}

In our ZSL model, the feature vector of each image is extracted by a pre-trained ResNet101 model, proposed by Xian {\it et al.} ~\cite{res101}. The normalized histogram of feature values is shown in Figure \ref{Fig:hist} and some statistical metrics given in Table \ref{Table:feature_dis}. As argued by Cacheux {\it et al.} ~\cite{tripletloss}, unbounded feature values may prevent Neural Network models from learning using triplet loss. 

One way to sidestep this problem is to simply bound the feature space. As shown in Table \ref{Table:feature_dis}, we found that 99.90\% of the feature values are below a threshold of about $7$ for each dataset. We thus propose to bound the feature space by preprocessing feature vectors for each dataset by clipping values above $7$ before normalizing to the range $[0,1]$. This is a simpler approach compared with the partial normalization approach proposed by Cacheux {\it et al.} ~\cite{tripletloss}.

\section{Ablation Study on hyperparameters}
We report more detailed results for the methods reported in the ablation study section in the main paper. In Table \ref{Table:clip_ablation}, we report the performance of our model influenced by different clipping values used in data preprocessing. As discussed in the previous section, the objective of clipping is to create a bounded feature space such that our Neural Network model can efficiently be trained using triplet loss. As shown in the table,  the performance of our model is better when using feature clipping than without feature clipping. 

In Table \ref{Table:delta_ablation}, we report the performance of our model with different $\Delta$ values in the triplet loss. {In our model, $\Delta$ is determined by a empirical gird search, with a coarse grid search in range $(0,100]$, followed by a fine grid search in range $(0.25,10]$.} $\Delta$ serves as a threshold in the triplet loss. A triplet $\{x^i_1,x^i_2,x^j_1\}$ is a trivial triplet if the inter-class pair distance exceeds the intra-class pair distance by a margin $\Delta$: $(x^i_1-x^j_1)^2 - (x^i_1-x^i_2)^2 > \Delta$. 

According to the triplet loss equation, trivial triplets will not have an influence on back-propagation gradients of the Neural Network. Small values for $\Delta$ may result in only a few non-trivial triplets, thus lowering the performance, while large $\Delta$ may add unnecessary computational cost when training the model.

As shown in Table \ref{Table:delta_ablation}, our model maintains a good performance with $\Delta\ge 3$ on all ZSL datasets. The performance peaks at $\Delta\approx 4$ and decreases slightly for larger $\Delta$ applied in the triplet loss $\mathcal{L}_{BT}$.

\section{Performance on Incorrect ``Proposed Split"}
{%The reproduced results reported in Table \ref{Table:performance} have been fine-tuned over ``Proposed Split V2.0" for better performance. However, some works did not provide their hyperparameter searching techniques, hence our fine-tuning may not give optimal performance.
To ensure a fair comparison, in Table \ref{Table:performance2}, we also compare our model's performance on the original ``Proposed Split" with results reported by previous SOTA papers, including f-VAEGAN-D2 ~\cite{f-VAEGAN-D2}, RELATION NET ~\cite{RelationNet}, DAZLE ~\cite{DAZLE}, Li {\it et al.} ~\cite{rethinking}, E-PGN ~\cite{EPGN}, OCD ~\cite{OCD}, DVBE ~\cite{DVBE}, TF-VAEGAN ~\cite{TF-VAEGAN}, IZF ~\cite{IZF}, AGZSL ~\cite{AGZSL}, IPN ~\cite{IPN} and CE-GZSL ~\cite{CE-GZSL}. We have not listed SOTA models that only report ImageNet performance like DGP ~\cite{DGP} and HVE ~\cite{HVE}, or only report transductive ZSL results like SDGN ~\cite{wu2020self}.}

{As can be seen from Table \ref{Table:performance2}, on ``Proposed Split", our model has reached SOTA performance on SUN and APY datasets. By comparing results shown in Table II in main paper and Table \ref{Table:performance2}, it can be seen that all previous reproduced works have a performance decrease after switching from ``Proposed Split" to the correct ``Proposed Split V2.0". On the contrary, although our model also reports a performance decrease on fine-grained datasets CUB and SUN, it maintains relatively stable performance on coarse-grained datasets AWA1, AWA2 and APY. This may due to the fact that our model has a simple structure and is less prone to overfitting.}

\begin{table*}[h!]
\centering
\scriptsize
%\tiny
\setlength\tabcolsep{2pt}
\begin{tabular}{c||c|c c c||c|c c c||c|c c c||c|c c c ||c|c c c} \hline
 \multirow{3}{*}{Clip Value} & \multicolumn{4}{c||}{CUB} &  \multicolumn{4}{c||}{SUN} & \multicolumn{4}{c||}{AWA2} & \multicolumn{4}{c||}{AWA1} & \multicolumn{4}{c}{APY} \\ \cline{2-21}
       & ZSL & \multicolumn{3}{|c||}{GZSL} & ZSL & \multicolumn{3}{|c||}{GZSL} & ZSL & \multicolumn{3}{|c||}{GZSL} & ZSL & \multicolumn{3}{|c||}{GZSL} & ZSL & \multicolumn{3}{|c}{GZSL}\\ \cline{2-21}
       & $A_T$ & $A_{U}$ & $A_{S} $ & $H$ & $A_T$ & $A_{U}$ & $A_{S}$ & $H$ & $A_T$ & $A_{U}$ & $A_{S}$ & $H$ & $A_T$ & $A_{U}$ & $A_{S}$ & $H$ & $A_T$ & $A_{U}$ & $A_{S}$ & $H$\\ \hline
3 & \textbf{61.2} & \textbf{52.0} & 57.2 & \textbf{54.5} & 59.7 & 48.9 & 33.1 & 39.5 & 67.8 & \textbf{63.5} & 75.6 & \textbf{69.0}& 67.4 & 62.6 & 73.1 & 67.5 & 40.2 & 32.9 & \textbf{72.7} & 45.3 \\
4 & 60.0 & 49.2 & \textbf{58.3} & 53.4 & 59.8 & 48.1 & 33.8 & 39.7 & 67.5 & 62.6 & 76.5 & 68.8 & 69.1 & 65.3 & 71.1 & 68.1 & 44.4 & 38.1 & 68.4 & 48.9 \\
5 & 59.5 & 49.1 & 57.3 & 52.9 & 61.4 & 49.3 & 33.6 & 40.0 & 67.8 & 62.1 & \textbf{77.2} & 68.8 & 69.7 & \textbf{66.0} & 70.6 & 68.2 & 45.5 & 40.0 & 67.0 & 50.1 \\
6 & 59.9 & 49.8 & 56.6 & 53.0 & 62.6 & 49.4 & \textbf{35.2} & \textbf{41.1} & 68.5 & 63.1 & 75.0 & 68.5 & 70.2 & 65.3 & 71.5 & 68.3 & 44.9 & 38.4 & 71.5 & 50.0 \\
7 & 60.1 & 50.3 & 56.0 & 53.0 & \textbf{63.2} & 50.4 & 34.8 & \textbf{41.1} & \textbf{68.6} & 62.2 & 76.6 & 68.7 & 70.0 & 64.5 & \textbf{73.3} & \textbf{68.6} & \textbf{47.1} & \textbf{42.8} & 64.3 & 51.4 \\
8 & 59.7 & 50.3 & 55.6 & 52.8 & 63.1 & 50.0 & 34.5 & 40.8 & 68.3 & 62.2 & 74.8 & 67.9 & 70.0 & 65.1 & 70.5 & 67.7 & 46.9 & 41.6 & 68.4 & 51.7 \\
9 & 60.0 & 48.6 & 57.8 & 52.8 & 62.8 & 50.0 & 34.5 & 40.8 & 68.0 & 61.2 & 75.3 & 67.5 & 69.9 & 63.8 & 72.6 & 67.9 & 46.1 & 42.4 & 62.3 & 50.5 \\
10 & 60.2 & 50.7 & 55.7 & 53.1 & 62.6 & 51.9 & 31.9 & 39.5 & 68.3 & 60.5 & 76.2 & 67.4 & \textbf{70.7} & 63.8 & 73.1 & 68.2 & 47.0 & 42.2 & 69.7 & \textbf{52.6} \\
None & 60.0 & 48.6 & 57.5 & 52.7 & 62.0 & 47.1 & 34.3 & 39.7 & 67.0 & 60.1 & 74.4 & 66.5 & 69.5 & 63.0 & 71.5 & 67.0 & 44.8 & 41.0 & 65.1 & 50.3 \\
\hline
\end{tabular}
\caption{Ablation Study with Clip number selected during feature preprocessing, with all other parts of the model fixed. Our model has a better performance with feature clipping than without feature clipping in data preprocessing. The performance of our model is robust when varying the clipping value around the proposed threshold 7. Moreover, different clip values have only a slight influence on our model's performance. }%\NB{Perhaps you should also say that the model is reasonably robust to small changes in the Clip number around the chosen value of 7.}
\setlength\tabcolsep{6pt}
\label{Table:clip_ablation}
\end{table*}

\begin{table*}[h!]
\centering
\scriptsize
%\tiny
\setlength\tabcolsep{2pt}
\begin{tabular}{c||c|c c c||c|c c c||c|c c c||c|c c c ||c|c c c} \hline
 \multirow{3}{*}{$\Delta$ Value} & \multicolumn{4}{c||}{CUB} &  \multicolumn{4}{c||}{SUN} & \multicolumn{4}{c||}{AWA2} & \multicolumn{4}{c||}{AWA1} & \multicolumn{4}{c}{APY} \\ \cline{2-21}
       & ZSL & \multicolumn{3}{|c||}{GZSL} & ZSL & \multicolumn{3}{|c||}{GZSL} & ZSL & \multicolumn{3}{|c||}{GZSL} & ZSL & \multicolumn{3}{|c||}{GZSL} & ZSL & \multicolumn{3}{|c}{GZSL}\\ \cline{2-21}
       & $A_T$ & $A_{U}$ & $A_{S} $ & $H$ & $A_T$ & $A_{U}$ & $A_{S}$ & $H$ & $A_T$ & $A_{U}$ & $A_{S}$ & $H$ & $A_T$ & $A_{U}$ & $A_{S}$ & $H$ & $A_T$ & $A_{U}$ & $A_{S}$ & $H$\\ \hline
0.25 & 58.2 & \textbf{57.1} & 33.2 & 42.0 & 60.3 & \textbf{56.2} & 22.6 & 32.3 & 66.2 & \textbf{65.0} & 48.0 & 55.2 & 69.3 & 68.3 & 49.5 & 57.4 & 40.3 & 38.5 & 37.9 & 38.2 \\
0.5 & 59.1 & 52.6 & 50.7 & 51.6 & 60.5 & 56.8 & 21.9 & 31.6 & 67.7 & 64.7 & 62.6 & 63.6 & 69.2 & 66.7 & 60.4 & 63.4 & 42.1 & 39.0 & 53.8 & 45.2 \\
1 & 59.3 & 53.9 & 49.9 & 51.8 & 61.3 & 54.9 & 28.2 & 37.3 & 67.2 & 61.2 & 74.5 & 67.2 & 69.2 & 63.4 & 71.9 & 67.4 & 44.7 & 40.2 & 61.3 & 48.5 \\
2 & 59.8 & 51.8 & 53.8 & 52.8 & 62.3 & 49.9 & 34.5 & 40.8 & 67.9 & 62.5 & 73.6 & 67.6 & 70.0 & 65.4 & 70.3 & 67.8 & 46.6 & 41.2 & 66.4 & 50.8 \\
3 & 59.9 & 50.7 & 55.1 & 52.8 & 62.6 & 50.6 & 34.6 & 41.1 & 68.5 & 62.3 & 75.4 & 68.2 & \textbf{70.2} & 64.7 & 72.5 & 68.4 & 45.3 & 40.6 & 65.8 & 50.2 \\
4 & \textbf{60.1} & 50.3 & 56.0 & \textbf{53.0} & 63.2 & 50.4 & 34.8 & 41.1 & 68.6 & 62.2 & 76.6 & \textbf{68.7} & 70.0 & 64.5 & \textbf{73.3} & \textbf{68.6} & \textbf{47.1} & \textbf{42.8} & 64.3 & \textbf{51.4} \\
5 & 59.9 & 49.6 & 56.2 & 52.7 & 63.2 & 50.3 & \textbf{35.0} & \textbf{41.3} & 68.9 & 61.9 & 76.6 & 68.5 & 69.5 & 64.0 & 72.6 & 68.0 & 45.9 & 41.9 & 62.6 & 50.2 \\
6 & 59.5 & 49.2 & 56.2 & 52.5 & 63.4 & 51.6 & 34.0 & 41.0 & \textbf{69.2} & 62.0 & 76.9 & 68.6 & 69.8 & \textbf{64.8} & 71.8 & 68.1 & 46.3 & 40.2 & \textbf{71.4} & \textbf{51.4} \\
7 & 59.5 & 48.8 & \textbf{56.5} & 52.4 & \textbf{63.5} & 51.1 & 34.5 & 41.2 & 68.7 & 62.0 & 76.6 & 68.5 & 70.1 & \textbf{64.8} & 72.3 & 68.3 & 44.5 & 40.2 & 66.2 & 50.0 \\
8 & 59.4 & 48.5 & \textbf{56.5} & 52.2 & 63.2 & 50.6 & 34.5 & 41.1 & 69.0 & 62.0 & 76.5 & 68.5 & 69.9 & 64.3 & 73.0 & 68.3 & 45.1 & 39.4 & 70.8 & 50.6 \\
\hline
\end{tabular}
\caption{Ablation Study with a threshold $\Delta$ in the triplet loss, with all the other parts of the model fixed. As long as $\Delta\ge 3$, our model has relatively stable performance.}
\setlength\tabcolsep{6pt}
\label{Table:delta_ablation}
\end{table*}

\begin{table*}[h!]
\centering
%\tiny
%\footnotesize
\scriptsize 
\setlength\tabcolsep{2pt}
\begin{tabular}{c||c|c c c||c|c c c||c|c c c||c|c c c ||c|c c c} \hline
 \multirow{3}{*}{Methods} & \multicolumn{4}{c||}{CUB} & \multicolumn{4}{c||}{SUN} & \multicolumn{4}{c||}{AWA2} & \multicolumn{4}{c||}{AWA1} & \multicolumn{4}{c}{APY} \\ \cline{2-21}
    & ZSL & \multicolumn{3}{|c||}{GZSL} & ZSL & \multicolumn{3}{|c||}{GZSL} & ZSL & \multicolumn{3}{|c||}{GZSL} & ZSL & \multicolumn{3}{|c||}{GZSL} & ZSL & \multicolumn{3}{|c}{GZSL}\\ \cline{2-21}
    & $A_T$ & $A_{U}$ & $A_{S} $ & $H$ & $A_T$ & $A_{U}$ & $A_{S}$ & $H$ & $A_T$ & $A_{U}$ & $A_{S}$ & $H$ & $A_T$ & $A_{U}$ & $A_{S}$ & $H$ & $A_T$ & $A_{U}$ & $A_{S}$ & $H$\\ \hline
SYNC ~\cite{SYNC} & 55.6 & 11.5 & 70.9 & 19.8 & 56.3 & 7.9 & 43.3 & 13.4 & 46.6 & 10.0 & 90.5 & 18.0 & 54.0 & 8.9 & 87.3 & 16.2 & 23.9 & 7.4 & 66.3 & 13.3 \\
GFZSL ~\cite{GFZSL} & 49.3 & 0.0 & 45.7 & 0.0 & 60.6 & 0.0 & 39.6 & 0.0 & 63.8 & 2.5 & 80.1 & 4.8 & 68.3 & 1.8 & 80.3 & 3.5 & 38.4 & 0.0 & 83.3 & 0.0\\
ALE ~\cite{ALE} & 54.9 & 23.7 & 62.8 & 34.4 & 58.1 & 21.8 & 33.1 & 26.3 & 62.5 & 14.0 & 81.8 & 23.9 & 59.9 & 16.8 & 76.1 & 27.5 & 39.7 & 4.7 & 73.6 & 8.7 \\
DEVISE ~\cite{DEVISE} & 52.0 & 23.8 & 53.0 & 32.8 & 56.5 & 16.9 & 27.4 & 20.9 & 59.7 & 17.1 & 74.7 & 27.8 & 54.2 & 13.4 & 68.7 & 22.4 & 39.8 & 4.9 & 76.9 & 9.2 \\
GDAN ~\cite{GDAN} & - & 39.3 & 66.7 & 49.5 & - & 38.1 & \textbf{89.9} & 53.4 & - & 32.1 & 67.5 & 43.5 & - & - & - & - & - & 30.4 & 75.0 & 43.4\\ 
CADA-VAE ~\cite{CADA-VAE} & - & 51.6 & 53.5 & 52.4 & - & 47.2 & 35.7 & 40.6 & - & 55.8 & 75.0 & 63.9 & - & 57.3 & 72.8 & 64.1 & - & - & - & - \\
TF-VAEGAN ~\cite{TF-VAEGAN} & 64.9 & 52.8 & 64.7 & 58.1 & 66.0 & 45.6 & 40.7 & 43.0 & - & - & - & - & 72.2 & 59.8 & 75.1 & 66.6 & - & - & - & -\\ 
f-VAEGAN-D2 ~\cite{f-VAEGAN-D2} & 61.0 & 48.4 & 60.1 & 53.6 & 65.6 & 50.1 & 37.8 & 43.1 & - & - & - & - & 71.1 & 57.6 & 70.6 & 63.5 & - & - & - & - \\
RELATION NET ~\cite{RelationNet} & 55.6 & 38.1 & 61.1 & 47.0 & - & - & - & - & 64.2 & 30.0 & \textbf{93.4} & 45.3 & 68.2 & 31.4 & \textbf{91.3} & 46.7 & - & - & - & - \\
DAZLE ~\cite{DAZLE} & - & 59.6 & 56.7 & 58.1 & - & 24.3 & 52.3 & 33.2 & - & 60.3 & 75.7 & 67.1 & - & - & - & - & - & - & - & - \\ 
Li {\it et al.} ~\cite{rethinking} & 54.4 & 47.4 & 47.6 & 47.5 & 60.8 & 42.6 & 36.6 & 39.4 & 71.1 & 56.4 & 81.4 & 66.7 & 70.9 & 62.7 & 77.0 & 69.1 & 38.0 & 26.5 & 74.0 & 39.0 \\
E-PGN ~\cite{EPGN}& \textbf{72.4} & 52.0 & 61.1 & 56.2 & - & - & - & - & 73.4 & 52.6 & 83.5 & 64.6 & \textbf{74.4} & 62.1 & 83.4 & \textbf{71.2} & - & - & - & - \\ 
DVBE ~\cite{DVBE} & - & 53.2 & 60.2 & 56.5 & - & 45.0 & 37.2 & 40.7 & - & \textbf{63.6} & 70.8 & 67.0 & - & - & - & - & - & 32.6 & 58.3 & 41.8 \\
OCD ~\cite{OCD} & - & 44.8 & 59.9 & 51.3 & - & 44.8 & 42.9 & 43.8 & - & 59.5 & 73.4 & 65.7 & - & - & - & - & - & - & - & - \\
%RFF-GZSL ~\cite{RFF}1-nn & - & 50.6 & 79.1 & 61.7 & - & 56.6 & 42.8 & 48.7 & - & - & - & - & - & 59.5 & 91.6 & 72.1 & - & - & - & -\\ 
%RFF-GZSL ~\cite{RFF}5-nn & - & \textbf{59.8} & \textbf{79.9} & \textbf{68.4} & - & \textbf{58.8} & 45.3 & 51.2 & - & - & - & - & - & \textbf{67.1} & \textbf{91.9} & \textbf{77.5} & - & - & - & -\\ 
IZF-Softmax ~\cite{IZF} & 67.1 & 52.7 & 68.0 & 59.4 & \textbf{68.4} & 52.7 & 57.0 & 54.8 & \textbf{74.5} & 60.6 & 77.5 & 68.0 & 74.3 & 61.3 & 80.5 & 69.6 & 44.9 & 42.3 & 60.5 & 49.8 \\ 
AGZSL ~\cite{AGZSL} & 57.2 & 41.4 & 49.7 & 45.2 & 63.3 & 29.9 & 40.2 & 34.3 & 73.8 & 65.1 & 78.9 & 71.3 & - & - & - & - & 41.0 & 35.1 & 65.5 & 45.7 \\
IPN ~\cite{IPN} & - & 60.2 & 73.8 & 66.3 & - & - & - & - & - & 67.5 & 79.2 & \textbf{72.9} & - & - & - & - & - & 37.2 & 66.0 & 47.6 \\
CE-GZSL ~\cite{CE-GZSL} & - & 63.9 & 66.8 & 65.3 & - & 48.8 & 38.6 & 43.1 & - & 63.1 & 78.6 & 70.0 & - & \textbf{65.3} & 73.4 & 69.1 & - & - & - & - \\
\hline 
\textbf{Ours} & 61.0 & 51.1 & 71.0 & 59.4 & 64.3 & \textbf{53.6} & 61.6 & \textbf{57.3} & 67.9 & 61.1 & 78.3 & 68.6 & 71.2 & 64.5 & 76.1 & \textbf{69.8} & \textbf{48.4} & \textbf{42.6} & \textbf{74.5} & \textbf{54.2}\\ 

\hline
\end{tabular}
\caption{Zero-Shot Learning Top-1 per-class Accuracy on incorrect ``Proposed Split". Results of each model are reported by original papers. Although our model is less prone to overfitting, we still outperforms previous papers on SUN and APY dataset. Some works are not included in Table II due to unavailable published official code.}
\setlength\tabcolsep{6pt}
\label{Table:performance2}
\end{table*}

\newpage
~\newpage
~\newpage
~\newpage
\bibliographystyle{./IEEEtran}
\bibliography{./ref.bib}